\documentclass[10pt,journal,compsoc]{IEEEtran}
\usepackage{url}
\usepackage{subfigure}
\usepackage{multirow}
\usepackage{graphicx}
\usepackage{booktabs}
\usepackage{amsmath}
\usepackage{amsfonts}
\usepackage{xcolor}
\usepackage{dsfont}
\ifCLASSOPTIONcompsoc
  \usepackage[nocompress]{cite}
\else
  \usepackage{cite}
\fi
\ifCLASSINFOpdf
\else
\fi
\begin{document}
\title{Weakly-supervised Micro- and Macro-expression Spotting Based on Multi-level Consistency}

\author{Wang-Wang~Yu,
        Kai-Fu Yang,
        Hong-Mei Yan,
        and~Yong-Jie Li,~\IEEEmembership{Senior Member,~IEEE}% <-this % stops a space
\IEEEcompsocitemizethanks{
\IEEEcompsocthanksitem W. Yu, K. Yang, H. Yan and Y. Li are with MOE Key Lab for NeuroInformation, 
University Of Electronic Science And Technology Of China, Chengdu,
China. Email: yuwangwang91@163.com, yangkf@uestc.edu.cn, hmyan@uestc.edu.cn, liyj@uestc.edu.cn\protect}% <-this % stops an unwanted space
\thanks{Manuscript received April 19, 2005; revised August 26, 2015. (Corresponding author: Yong-Jie Li, Kai-Fu Yang.)}}


\markboth{Journal of \LaTeX\ Class Files,~Vol.~14, No.~8, August~2015}%
{Shell \MakeLowercase{\textit{et al.}}: Bare Demo of IEEEtran.cls for Computer Society Journals}



\IEEEtitleabstractindextext{%
\begin{abstract}
Most micro- and macro-expression spotting methods in untrimmed videos suffer from the burden of video-wise collection and frame-wise annotation. 
Weakly-supervised expression spotting (WES) based on video-level labels can potentially mitigate the complexity of frame-level annotation while 
achieving fine-grained frame-level spotting. However, we argue that existing weakly-supervised methods are based on multiple instance learning (MIL) involving 
inter-modality, inter-sample, and inter-task gaps. The inter-sample gap is primarily from the sample distribution and duration. Therefore, we propose a novel 
and simple WES framework, MC-WES, using multi-consistency collaborative mechanisms that include modal-level saliency, video-level distribution, label-level duration 
and segment-level feature consistency strategies to implement fine frame-level spotting with only video-level labels to alleviate the above gaps and merge 
prior knowledge. The modal-level saliency consistency strategy focuses on capturing key correlations between raw images and optical flow. The video-level 
distribution consistency strategy utilizes the difference of sparsity in temporal distribution. The label-level duration consistency strategy exploits the 
difference in the duration of facial muscles. The segment-level feature consistency strategy emphasizes that features under the same labels maintain similarity. 
Experimental results on three challenging datasets--CAS(ME)$^2$, \textcolor{black}{CAS(ME)$^3$}, and SAMM-LV--demonstrate that MC-WES is comparable to state-of-the-art fully-supervised methods.
\end{abstract}

}

\maketitle

\IEEEdisplaynontitleabstractindextext

% \fi
\IEEEpeerreviewmaketitle

\IEEEraisesectionheading{\section{Introduction}\label{sec:introduction}}


\IEEEPARstart{F}{acial} expression is an important medium for conveying human emotions. Expressions can be categorized as micro-expressions (MEs) and
macro-expressions (MaEs) \cite{ekman1969nonverbal}. MEs are subtle, involuntary facial movements and often occur when a person tries to conceal or suppress 
his or her true emotion. MEs contain three prominent features on the face--short duration, low intensity, and local movement \cite{wang2021mesnet}--making them
difficult even for experienced experts to recognize \cite{frank2009see}. In contrast, MaEs are visible facial motion processes with distinct start and end 
temporal points and variable durations. Compared with MaEs, which may convey inauthentic emotions, MEs reflect real changes in emotion, and are therefore useful 
in high-stakes environments such as medical diagnosis, public safety, crime investigation and political business negotiation \cite{ekman2003darwin, ekman2009lie}.

\begin{figure}[tbp]
  \centering
  \includegraphics[width=\linewidth]{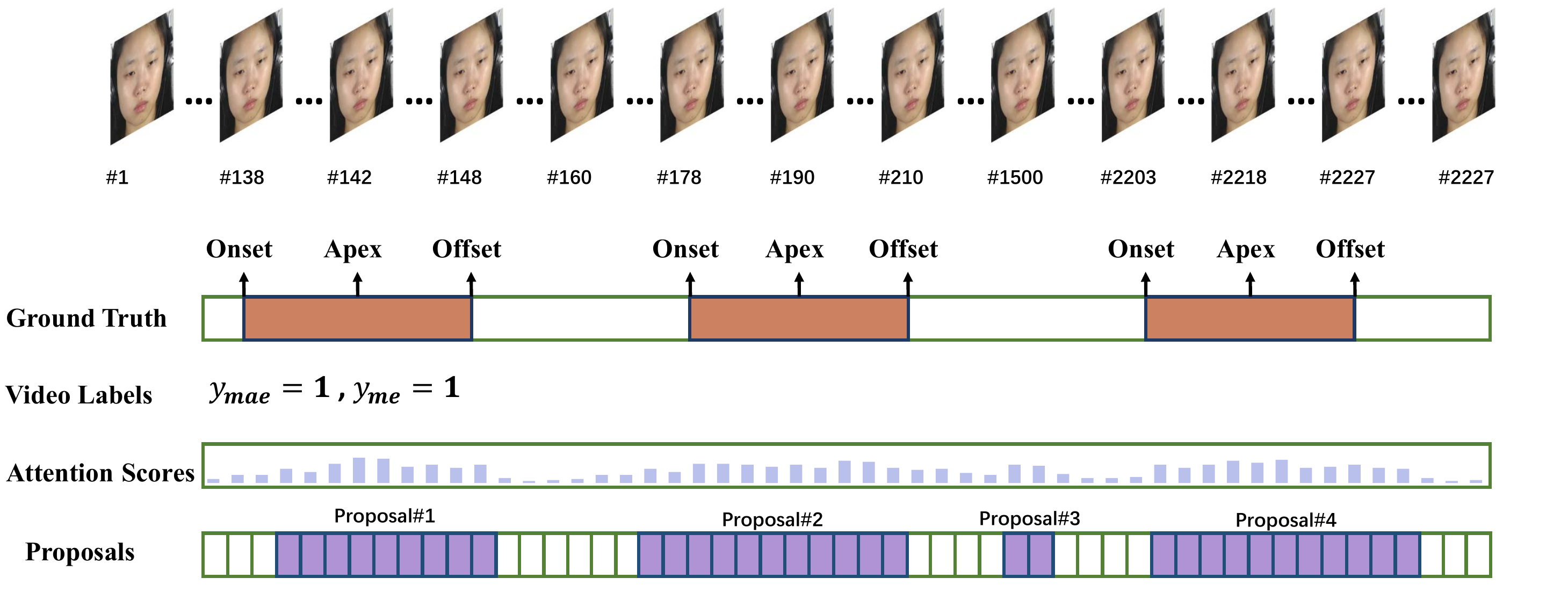}
  \caption{A video of S15\_0502 with the range of frames \#1 to \#2273 from the CAS(ME)$^2$ dataset. There are three ground truth intervals, marked with specific onset, 
  apex and offset frames. The first interval contains a micro-expression, and the last two contain macro-expressions.
  \textcolor{black}{Therefore, video-level labels of this video contain both micro-expression and macro-expression classes, i.e., $y_{me}=1$ and $y_{mae}=1$.}
  In the pre-processing phase, the video is divided into a series of uniform non-overlapping snippets. During training, we generate attention scores indicating actionness 
  scores (probabilities belonging to foreground) with the video-level labels. During testing, these attention scores are used to generate multiple proposals (e.g., the 
  four proposals filled in purple) with different top-$k$ values. The goal is to spot several consecutive video snippets as close as possible to ground truth intervals.}
  \label{fig1}
\end{figure}  
Expression analysis includes the major tasks of spotting and recognition. Recognition aims to identify facial expressions as belonging to specific emotional 
categories \cite{xie2019adaptive} or continuous multidimensional values \cite{du2014compound, li2017reliable, liang2020fine}. 
Spotting, as a prior task, focuses on localizing key continuous intervals from an untrimmed long video and classifying them as MEs and MaEs. 
As it is difficult to quantify the intensity and range of the movement of expressions \cite{lu2022more}, the duration of MaEs and MEs is naturally seen as a benchmark for 
classification on all main spotting datasets, including CAS(ME)$^2$ \cite{qu2017cas}, SAMM-LV \cite{yap2020samm}, MMEW \cite{ben2021video}, and CAS(ME)$^3$ \cite{li2022cas}.
This practice is based on the statistical observation that MaEs typically last between 0.5 to 4.0 seconds, while MEs occur in less than 0.5 second \cite{ekman2003darwin}.
To portray the whole change of facial movements with a more fine-grained description, an expression can be described by three key temporal points--onset, apex, and 
offset \cite{wang2021mesnet}--as illustrated in Figure \ref{fig1}. The onset is the starting time, the apex demonstrates the most noticeable emotional information under 
maximum facial muscle deformation \cite{ekman1993facial, esposito2007amount}, and the offset is the ending time. From this perspective, datasets for 
the spotting task furnish the onset, offset, and apex frames of all ground truths for model learning.

ME and MaE spotting have been shown to be successful in long untrimmed videos based on frame-by-frame annotations in a fully-supervised setting \cite{he2022micro, 
yu2021lssnet}. However, extensive video-level acquisition relies heavily on carefully designed experimental environments and stimulus conditions \cite{qu2017cas, 
yap2020samm, ben2021video, li2022cas}. Moreover, obtaining fine-grained frame-level labels requires extensive manual labor involving two or more coders, with an 
average of two hours required to annotate one minute of ME and MaE videos \cite{bartlett2010insights}. These bring difficulty in rolling out ME-related applications 
on a large scale. 

There is a growing availability of diverse face videos with emotion labels on the internet, providing a potential source for data collection. Although such video-level 
labels are essentially weak labels, they provide direct emotional clues. This motivates us to develop an effective weak label-based ME and MaE spotting method. Our goal 
is to achieve automatic weakly-supervised expression spotting (WES) with only video-level (weak) labels, as illustrated in Figure \ref{fig1}. 
Obtaining these video-level labels, however, presents a challenge. Most spotting datasets \cite{qu2017cas, ben2021video, li2022cas} rely on labeling action units (AUs)
to determine the onset and offset frames of ground truth intervals. Then coders classify the labeled intervals into MEs and MaEs based on their durations. 
To significantly reduce the time and manual effort required for annotation, 
\textcolor{black}{the proposed method only requires two labels for each video: (1) whether there is an ME, and (2) whether there is an MaE.} 

\textcolor{black}{The WES framework is basically based on multiple instance learning (MIL) \cite{Dmaron1997framework}, which is a type of machine learning where bags of instances 
are classified instead of individual items. Therefore, this requires us to construct a series of positive and negative bags.} To this end, the videos are divided into uniform non-overlapping 
snippets \footnote{In this paper, we treat snippets as the smallest granularity, and intervals or proposals as sequences consisting of one or more consecutive snippets.} as 
instances. The essence of WES task is that, during training, only the expression categories (ME, MaE) contained in the video are given for generating bags, but 
not the number of expressions and the onset and offset frames. During testing, this task involves the localization and classification of expression intervals by 
computing differences between snippets and combining continuous snippets to create proposals.

\begin{figure*}[tbp]
  \centering
  \includegraphics[width=\linewidth]{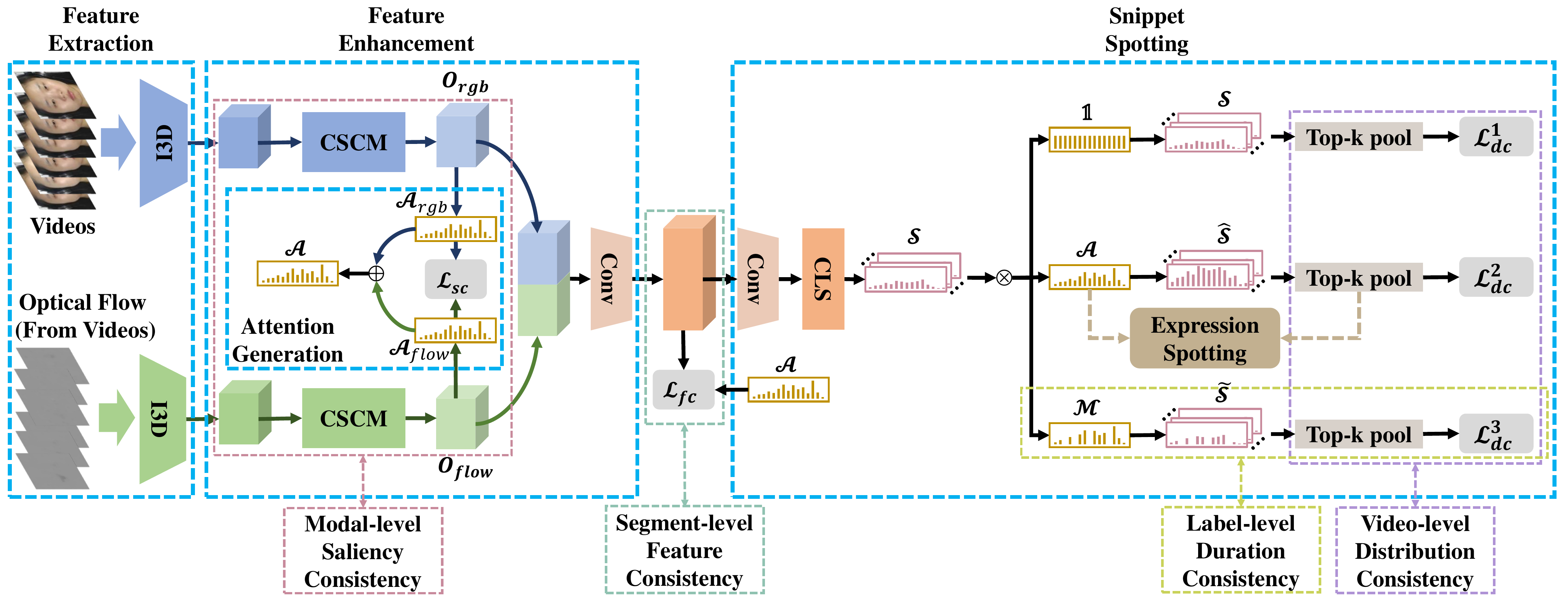}
  \caption{Overall schematic of proposed MC-WES. 
  Given a video, snippet-level features are extracted by the two-stream Inflated 3D ConvNets (I3D) model \cite{carreira2017quo} from a set of uniform non-overlapping snippets 
  sampled from videos and optical flow. CSCM is the core saliency compensation module used to fuse core features and filter irrelevant features for each modality. Processed 
  features (i.e, $O_{rgb}$ and $O_{flow}$) are used to generate attention scores which are used to calculate modal-level saliency consistency loss $\mathcal{L}_{sc}$. 
  Mean attention scores $\mathcal{A}$ of two modalities represent probability that snippets belong to the foreground. 
  \textcolor{black}{These processed features are also concatenated and fed into a convolution layer to produce modal-enhanced features, which are used to implement the segment-level feature 
  consistency strategy. This strategy is then used to optimize features corresponding to top-$k$ attention scores with the attention-guided feature consistency loss $\mathcal{L}_{fc}$.} 
  Modal-enhanced features are subsequently fed into classifier $CLS$ to generate temporal class activation maps (T-CAMs) $\mathcal{S}$, which indicate logits for each category of video. 
  T-CAMs $\mathcal{S}$ are used to fuse with all-1 matrix $\mathds{1}$, attention scores $\mathcal{A}$, and strategy-filtered mask matrix $\mathcal{M}$, respectively. \textcolor{black}{Three branches of 
  fused T-CAMs $\mathcal{S}$ are processed with the video-level distribution consistency strategy, where we utilize temporal top-$K$ pooling layers to aggregate logits and calculate 
  MIL-based losses (i.e., $\mathcal{L}_{dc}^1, \mathcal{L}_{dc}^2, \mathcal{L}_{dc}^3$).} Specifically, in third branch, we implement the label-level duration consistency strategy to 
  calculate the variation and generate the mask matrix $\mathcal{M}$ in attention scores between snippets within a certain range. \textcolor{black}{Then we implement attention-guided duration 
  consistency loss $\mathcal{L}_{dc}^3$ to highlight difference in duration between MEs and MaEs.} During testing, we utilize T-CAMs $\widehat{\mathcal{S}} $ and mean attention 
  scores $\mathcal{A}$ to generate expression proposals.}
  \label{fig2}
\end{figure*}  

To date, several MIL-based methods \cite{hong2021cross, islam2021hybrid, paul2018w} have been proposed for the weakly-supervised temporal action localization (WTAL) task. However, 
when we try to apply the MIL-based method directly to the WES framework, \textcolor{black}{inter-modal}, inter-sample and inter-task gaps are produced. The inter-modal gap occurs due to the use 
of features from two modalities, raw images and optical flow, as input in two-stream networks \cite{hong2021cross, islam2021hybrid, paul2018w, he2022asm}. Although optical flow 
can provide enough motion information and raw images can provide enough appearance information \cite{simonyan2014two, sevilla2018integration}, the features from the two modalities 
are inconsistent \cite{li2022forcing}. The inter-sample gaps are primarily manifested in sample distribution and duration. Specifically, the distribution reflecting the frequency 
of sample appearance is not uniform because MEs are more dependent on harsh excitation conditions than MaEs \cite{qu2017cas, yap2020samm}. \textcolor{black}{In addition, the duration varies for 
different expressions due to their different definitions \cite{paul2007emotions}.} The inter-task gap refers to the discrepancy between the localization 
and classification tasks \cite{liu2019completeness, liu2021weakly, liu2021acsnet}. \textcolor{black}{Existing action localization models \cite{fu2022compact, lee2021learning, huang2022weakly} supervised 
with video-level information tend to favor the most discriminative snippets or the contextual background, which may lead to localize inaccurate action boundaries or incorrect action snippets.}

To mitigate the above multiple gaps and merge more prior knowledge in weakly-supervised frameworks, we propose a framework, MC-WES, which employs the collaboration of multi-level 
consistency to spot more fine-grained expression intervals with video-level labels. The WC-WES framework includes four consistency strategies: (i) modal-level saliency consistency; 
(ii) video-level distribution consistency; (iii) label-level duration consistency; (iv) segment-level feature consistency. \textcolor{black}{In particular, the modal-level saliency consistency strategy 
is introduced to capture the significant correlations between the two modalities of raw images and optical flow. The video-level distribution consistency strategy is designed to incorporate
the prior knowledge about the different distributions of MEs and MaEs. The label-level duration consistency strategy aims to take into account the duration difference between MEs and MaEs. 
The segment-level feature consistency strategy is designed to minimize intra-class differences and maximize inter-class differences.}

We summarize our main contributions as follows:
\begin{itemize}
  \item To the best of our knowledge, this is the first work to utilize a weakly-supervised MIL-based learning framework for ME and MaE spotting in untrimmed face videos with only 
  coarse video-level labels and no fine-grained frame-level annotations.
  \item \textcolor{black}{A novel saliency compensation module (CSCM) is designed to extract effective and complementary features from the two modalities of raw images and optical flow. 
  CSCM works not only to remove redundant information, but also to extract salient information and enhance the information of both modalities.}
  \item \textcolor{black}{A modal-level saliency consistency strategy is proposed to address the information redundancy and alleviate the inter-modal asynchronization between the two modalities. 
  In particular, this strategy is realized by generating the modal-specific attention scores based the CSCM extracted features and guiding the model training with a modal-level saliency consistency loss.} 
  \item \textcolor{black}{To merge the prior knowledge about the sample distribution and duration between MEs and MaEs, a video-level distribution consistency strategy and a label-level duration 
  consistency strategy are respectively designed. Specifically, The former strategy involves selecting the snippet-level logits of different categories from each video using top-k pooling and 
  evaluating an attention-guided video-level distribution consistency loss, and the latter strategy involves removing potential ME snippets and computing an attention-guided duration consistency loss.}
  \item \textcolor{black}{A segment-level feature consistency strategy is proposed to highlight the similarity of features within the same categories in a video pair, which utilizes the top-k 
  localization-related attention scores to select classification-related features and logits and then evaluate an attention-guided feature consistency loss.}
  \item \textcolor{black}{Extensive experiments on the commonly used CAS(ME)$^2$, CAS(ME)$^3$, and SAMM-LV datasets demonstrate that our weakly-supervised MC-WES framework is comparable to existing 
  fully-supervised methods in terms of multiple metrics and clearly exceeds current common weakly-supervised methods.}
  \end{itemize}

% file is intended to serve as a ``starter file''
% for IEEE Computer Society journal papers produced under \LaTeX\ using
% IEEEtran.cls version 1.8b and later.
% I wish you the best of success.

 
\section{Related Work}

\subsection{Fully-supervised Expression Spotting}
According to the type of the generated proposals, fully-supervised spotting methods based on deep learning can be classified as either key frame- or interval-based. 
Key frame-based methods \cite{pan2020local, zhang2018smeconvnet, yap20213dcnn, liong2021shallow} aim to localize expression intervals by looking through one or more frames in a 
long video. Pan et al. \cite{pan2020local} identify each frame as MaE, ME, or background frame, which results in the loss of positive samples. In contrast, SMEConvNet \cite{zhang2018smeconvnet} 
uses one frame to spot intervals. However, most proposals generated by SMEConvNet tend to be of short duration. Yap et al. \cite{yap20213dcnn} rely on a few fixed durations, which 
may generate a large number of negative samples. SOFTNet \cite{liong2021shallow} uses a shallow optical flow three-stream convolutional neural network (CNN) to predict whether each 
frame belongs to an expression, and introduces pseudo-labeling to facilitate the learning process. 

Interval-based approaches \cite{sun2019two, tran2019dense, verburg2019micro, wang2021mesnet, yu2021lssnet} take all the image features in the video as input, and pay attention to 
the information of neighboring frames. To this end, long short-term memory (LSTM) is commonly used to encode neighbor temporal information \cite{sun2019two, tran2019dense, 
verburg2019micro}. However, LSTM cannot handle longer and more detailed temporal information. Wang et al. \cite{wang2021mesnet} utilize a clip proposal network to initially build 
long-range temporal dependencies by combining different scales and types of convolution layers with downsampled features. LSSNet \cite{yu2021lssnet} uses anchor-based and 
anchor-free branches to generate multi-scale proposal intervals based on snippet-level features from raw images and optical flow. 

Fully-supervised expression spotting methods generally achieve good performance, but they rely heavily on frame-level annotation which greatly increases the cost of 
data labeling. In contrast, weakly-supervised spotting methods try to only use the relatively effortless video-level labels to achieve frame-level localization.

\subsection{Weakly-supervised Temporal Action Localization}
Utilizing weak labels to train models has made significant progress in computer vision such as semantic segmentation \cite{zhang2020survey, ge2018multi, vezhnevets2011weakly}, object 
detection \cite{cheng2020high, bilen2015weakly}, and temporal action localization (TAL) \cite{islam2021hybrid, hong2021cross, paul2018w}. In contrast to the fully-supervised 
TAL \cite{lin2017single, yang2020revisiting, zhao2017temporal, zhang2020asfd}, the WTAL methods are free of extensive frame-level annotations and adopt video-
\cite{wang2017untrimmednets, nguyen2018weakly, nguyen2019weakly, paul2018w, su2018cascaded} or point (key frame)-level \cite{lee2021learning, ju2020point, ju2021divide, ma2020sf} 
labels during training. Since different video-level WTAL approaches have different emphases, we can categorize them as foreground-only, background-assisted or pseudo-label-guided.

Foreground-only WTAL methods focus on extracting effective foreground information. UntrimmedNet \cite{wang2017untrimmednets} introduces MIL, where treats snippets as separate 
instances, which are also used in selection and aggregation to obtain proposals. Later, STPN \cite{nguyen2018weakly} adds temporal class activation maps (T-CAMs) to generate 
one-dimensional temporal attentions, and aggregates proposals by adaptive temporal pooling operation. W-TALC \cite{paul2018w} utilizes a co-activity similarity loss in the 
video pairs to enhance the similarity of identically labeled snippets and the variability of differently labeled snippets. To integrate multi-scale temporal information, 
CPMN \cite{su2018cascaded} uses a cascaded pyramid mining network. 3C-Net \cite{narayan20193c} employs center and counting losses to learn more discriminative action features. 

\begin{figure}[tbp]
  \centering
  \includegraphics[width=\linewidth]{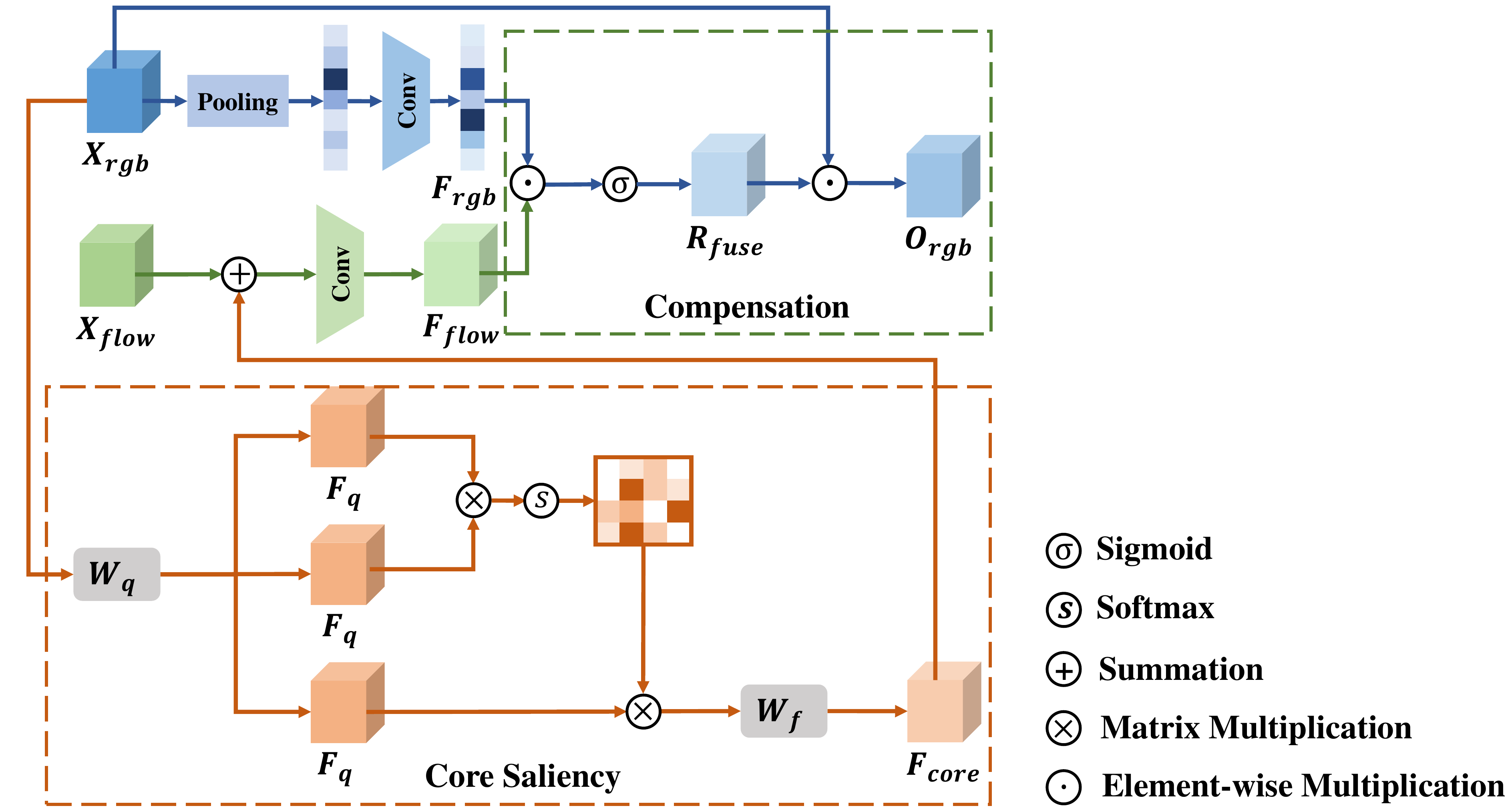}
  \caption{Core Saliency Compensation Module (CSCM).
  We utilize the raw image or optical flow modality as the main branch and the other as the supporting branch. The main branch is used to extract core saliency 
  (long-range element-wise dependency) information $F_{core}$ and element-wise squeezer $F_{rgb}$. Core saliency unit is part of self-attention mechanisms 
  \cite{vaswani2017attention}, with only one $1\times 1$ convolution layer $W_q$. Output $F_{core}$ from core saliency unit complements significant information and 
  reinforces the modal coherence information between two modalities in the supporting branch. The enhanced supporting branch cooperates with element-wise squeezer $F_{rgb}$ to 
  generate compensation unit $R_{fuse}$, which is modulated by a sigmoid function. Features from the main branch are enhanced by modulated $R_{fuse}$ as final output $O_{rgb}$.
  }
  \label{fig3}
\end{figure}  

Foreground-only WTAL methods do not take background frames as a separately guided class during training, although there are contexts in them associated with actions.
\textcolor{black}{For example, DGAM \cite{shi2020weakly} divides a ``longjump'' action into approaching, jumping and landing stages, in which preparing and finishing are the most crucial contexts.} 
To model the entire action completely, background-assisted methods build multi-branch or multi-stage architectures. CMCS \cite{liu2019completeness} uses a diversity loss to 
model integral actions, and a hard negative generation module to separate contexts. BaSNet \cite{lee2020background} adopts an attention branch to suppress
background interference. DGAM \cite{shi2020weakly} utilizes a two-stage conditional variational auto-encoder (VAE) \cite{kingma2013auto} to separate action and context frames.  
\textcolor{black}{In contrast, HAM-Net \cite{islam2021hybrid} models an action as a whole based on attention scores consisting of soft, semi-soft and hard attention.}

To minimize the discrepancy between classification and localization, current methods generally generate pseudo labels during training. RefineLoc \cite{pardo2021refineloc} iteratively 
generates pseudo labels that are used as supervised information to refine predictions for the next iteration. EM-MIL \cite{luo2020weakly} uses an expectation maximization 
algorithm to improve pseudo-label generation. ASM-Loc \cite{he2022asm} introduces a re-weighting module for pseudo label noise effects with an uncertainty prediction 
module \cite{nguyen2019weakly}. EM-Att \cite{huang2022weakly} mines the discriminative snippets and propagates information between snippets to generate pseudo labels.

Compared with video-level WTAL methods, point-level WTAL methods add a small amount of supervised information to localize more accurate action boundaries. Moltisanti et 
al. \cite{moltisanti2019action} introduce the point-level labels to bridge the gap between the growing variety of actions and weak video-level labels. SF-Net \cite{ma2020sf} 
adopts a pseudo label mining strategy to acquire more labeled frames. LACP \cite{lee2021learning} takes the points to search for optimal sequences, which are used to learn 
completeness of entire actions. Ju et al. \cite{ju2021divide} divide entire video into multiple video clips, and use a two-stage network to localize the action instances in each one.

In this study, we choose the video-level framework, even though point-level methods have been shown to produce better results, primarily because the intensity and duration of the expressions are 
weaker and shorter than those of the general actions in WTAL. This makes point-level annotation of expressions significantly more expensive. \textcolor{black}{Additionally, to enhance our model's 
ability to capture fine-grained and precise information, we also combine the ideas of background-assisted and pseudo-labeling methods when building our MC-WES framework.}

\subsection{Multiple Instance Learning (MIL)}
In a weakly-supervised learning framework using coarse-grained labels, MIL \cite{Dmaron1997framework} has demonstrated its effectiveness. \textcolor{black}{MIL involves creating batches 
of positive bags, each containing at least one positive sample, and negative bags containing no positive samples. The objective of MIL is to train a model to be capable 
of accurately predicting the labels of unseen bags.} Several tasks, including semantic segmentation and object detection, utilize region proposal techniques to generate
these bags \cite{zhang2020survey, ge2018multi, vezhnevets2011weakly, cheng2020high, bilen2015weakly}. In WTAL, most methods sample a large number of snippets from each 
video to construct the bags \cite{islam2021hybrid, hong2021cross, paul2018w}.

Because current action localization datasets \cite{idrees2017thumos, heilbron2015activitynet} commonly have more than 20 action categories, MIL-based methods 
\cite{islam2021hybrid, hong2021cross, paul2018w} do not focus on the differences between categories. In contrast, the spotting task in this paper involves only the categories 
of MEs and MaEs, allowing us to integrate their differences for more precise localization. Specifically, we work on distribution and duration without focusing on the intensity 
and range of face movements in our MC-WES framework. To achieve this, we incorporate duration information into the attention scores to process T-CAMs using multiple branches, 
and aggregate processed T-CAMs using the different values in the top-$K$ based on distribution information to calculate our MIL-based losses.

\section{Methodology}
We present a novel and simple weakly-supervised ME and MaE spotting framework, MC-WES, with only video-level weak labels to complete frame-level expression spotting. 
\subsection{Problem Formulation}
\label{Section 3.1}
Assume a video $V=\{{v_t}\}_{t=1}^L$ has $L$ frames and contains $n$ expression categories with video-wise labels $y_c\in \{ 1,0 \}^{n+1}$, where $n+1$ is the number 
of expression categories that contain the background class. $y_i=1$ means that there is at least one instance of the $i$-th expression, and if $y_i=0$, there is no such instance. 
In WES framework, the number and the order of expressions in the video are not provided in the training phase. During testing, expression 
proposals ${E} = (f_{on}, f_{off}, y, \phi)$ are generated, where $f_{on}$ denotes the onset frame, $f_{off}$ denotes the offset frame, $y$ denotes the category, and $\phi$ indicates 
the confidence score. Note that $f_{on}$ and $f_{off}$ are integer multiples of the number of frames in the snippets, because we only localize specific snippets to generate 
proposal intervals. Following a previous approach \cite{yu2021lssnet}, we only need to filter the samples with confidence scores and then calculate the recall and 
precision based on the duration of the proposals, defining those below 0.5 second as MEs and those longer than that as MaEs instead of the classification results 
of the model.

\subsection{Framework Overview}
\label{Section 3.2}
As shown in Figure \ref{fig2}, our MC-WES framework has four components: feature extraction, feature enhancement, attention generation, and snippet spotting. 

\noindent \textbf{Feature Extraction.}
We first generate optical flow from a video and then divide the video and optical flow into a series of uniform non-overlapping snippets, each containing $g$ frames.
\textcolor{black}{These snippets are used to extract image features $X_{rgb} \in \mathbb{R} ^{T\times D}$ and optical flow features $X_{flow} \in \mathbb{R} ^{T\times D}$ by 
the two-stream Inflated 3D ConvNets (I3D) model \cite{carreira2017quo}, where $T=L/g$ is the number of snippets, and $D$ is the dimension of one snippet feature. }
To maintain the consistency of the numbers of raw images and optical flow, we delete the last frame of a video. \textcolor{black}{The TV-L1 optical flow algorithm \cite{wedel2009improved} 
with the default smoothing parameter ($\lambda=0.15$) is used to generate a dense optical flow between adjacent frames.}

\noindent \textbf{Feature Enhancement.}
To incorporate movement and appearance information, we take the features from the raw image and optical flow modalities as input following previous frameworks 
\cite{hong2021cross, islam2021hybrid, lee2020background}. However, our input features are extracted from the I3D model used for action recognition, leading to 
feature redundancy \cite{hong2021cross}. In addition, the features are not synchronized due to the difference in modalities \cite{li2022forcing}. \textcolor{black}{To mitigate 
the feature redundancy and the differences across modalities, we enhance the features of each of the two modalities by task-specific feature complementary with the proposed Core 
Saliency Compensation Modules (CSCMs).}

\noindent \textbf{Attention Generation.}
Attention scores represent the probability that each snippet belongs to the foreground in our MC-WES framework. We compute modal-specific temporal attention scores 
based on the enhanced modal-specific features as $\mathcal{A}_{rgb}\in \mathbb{R} ^{T}$ for image modality and $\mathcal{A}_{flow}\in \mathbb{R} ^{T}$ for optical 
flow modality \cite{hong2021cross}. Then we calculate the mean class-agnostic attention scores $\mathcal{A} = \frac{1}{2}  (\mathcal{A}_{rgb} + \mathcal{A}_{flow})$ 
as guidance to process the \textcolor{black}{class-specific} logits of T-CAMs, which are generated by the following classifier. These mean attention scores are also used to select 
class-agnostic expression proposals during testing.

\noindent \textbf{Snippet Spotting.}
Snippet spotting is used to optimize class-specific T-CAMs $\mathcal{S} \in \mathbb{R} ^{T\times (n+1)}$, which indicate the logits of each snippet belonging to all categories \cite{shou2018autoloc}. \textcolor{black}{Here we use logits to represent the eigenvalues of MaE, ME, and the background before being processed by the softmax function. For instance, logits signify the class-specific eigenvalues prior to processing through the softmax 
function in a classification task.} Note that the $(n+1)$-th class is the background class. As shown in 
Figure \ref{fig2}, three branches process the logits of T-CAMs, two of which are coupled with the localization information derived from temporal attention scores.

\subsection{Multi-level Consistency Analysis}
\label{Section 3.3}
Expression spotting is the temporal localization and binary classification task in untrimmed face videos, and is essentially an application of TAL in expression analysis. 
Due to the success of WTAL in video understanding, we apply it to expression analysis and introduce a WES framework. As shown in Figure \ref{fig2}, to fuse more prior information 
and alleviate existing gaps, including inter-modal, inter-sample and inter-task, we employ a multi-consistency collaborative mechanisms, with four consistency strategies: 
modal-level saliency, video-level distribution, label-level duration and segment-level feature consistency. In particular, the modal-level saliency consistency strategy 
is introduced in feature enhancement to mitigate inter-modal gaps. The video-level distribution and label-level duration consistency strategies are used in snippet spotting 
to alleviate inter-sample gaps that arise from differences in distribution and duration. The segment-level feature consistency strategy is utilized for fused features to
mitigate inter-task gaps, which are intermediate between the components of feature enhancement and snippet spotting.

\noindent \textbf{Modal-level Saliency Consistency Strategy.}
Following previous frameworks \cite{hong2021cross, paul2018w, narayan20193c, islam2021hybrid, lee2020background}, we perform our MC-WES framework using 
a two-stream network, and extract features from the original images and optical flow as input using the I3D model \cite{carreira2017quo}. However, the 
I3D model is originally trained for video action recognition, and its extracted features always contain noisy information, which can degrade performance 
and lead to suboptimal training \cite{feng2021mist, lei2021less}. CO$_2$-Net \cite{hong2021cross} adopts cross-modal consensus modules (CCMs) to reduce task-irrelevant 
information redundancy. The process is a squeeze-and-excitation block from SENet \cite{hu2018squeeze}. However, the distributions of features from raw image and optical 
flow are temporally inconsistent \cite{li2022forcing}. As a result, the channel-wise descriptors generated by CCM tend to weaken modal-inconsistent information and strengthen 
some modal-consistent irrelevant information. \textcolor{black}{Inspired by these works, we design a novel CSCM to extract task-specific features based on complementary 
enhanced features by encouraging the complementarity of global core information between raw image and optical flow modalities. Its purpose is to capture significant correlations 
between the two modalities, mitigate suboptimalities resulting from their differences, and harness the strengths of each.}

To alleviate the information discrepancy caused by modality inconsistency, we use CSCM to extract the global core salient information of the main modality (i.e., raw images), 
which is used to enhance the auxiliary modality (i.e., optical flow). Then the enhanced auxiliary modality is integrated with the pooled 
elements of the main modality. Suppose the input features of raw images and optical flow are $X_{rgb} \in \mathbb{R} ^{T\times D}$ 
and $X_{flow} \in \mathbb{R} ^{T\times D}$, respectively, as shown in Figure \ref{fig3}. The input features of raw images are squeezed by an adaptive average 
pooling layer as a modal-specific global vector, which is used to aggregate information,
\begin{equation}
  F_{rgb} = \sigma (W_{rgb} \cdot (\frac{\sum_{t = 1}^{T} (X_{rgb})}{T} ) + B_{rgb}),
  \label{equ3.1}
\end{equation}
where $W_{rgb}$ is a $3\times 3$ convolution layer, $B_{rgb} \in \mathbb{R}^{D}$ is the bias of convolution layer $W^{rgb}$, and $\sigma$ is 
a sigmoid function to ensure the generated weights are between 0 and 1. We also take simple self-attention mechanisms \cite{vaswani2017attention} with
only one $1\times 1$ convolution layer and without a shortcut operation to process the input features of raw images to generate core salient information,
\begin{equation}
  F_q = W_{q} \cdot X_{rgb} + B_{q},
  \label{equ3.2}
\end{equation}
\begin{equation}
  F_{core} = W_{f} \cdot (\varepsilon (F_q \cdot F_q^T) \cdot F_q) + B_{f},
  \label{equ3.3}
\end{equation}
where $W_{f}$ and $W_{q}$ are $1\times 1$ convolution layers with respective biases $B_{f}$ and $B_{q}$, and $\varepsilon$ is a softmax function.
Then the core saliency $F_{core}$ is used to supplement information to features $X_{flow}$, whose output is
\begin{equation}
  F_{flow} = W_{flow} \cdot (X_{flow} + F_{core}) + B_{flow},
  \label{equ3.4}
\end{equation}
where $W_{flow}$ is a $3\times 3$ convolution layer with bias $B_{flow}$. 
We fuse $F_{flow}$ and $F_{rgb}$ to generate the channel-wise descriptor $R_{fuse}$ with a sigmoid function,
\begin{equation}
  R_{fuse}= \sigma (F_{flow} \odot F_{rgb}),
  \label{equ3.5}
\end{equation}
where ``$ \odot $'' denotes element-wise multiplication. 
Therefore, when the main modality is raw images, the output features $O_{rgb}$ of CSCM are defined as
\begin{equation}
  O_{rgb} = R_{fuse} \odot X_{rgb}.
  \label{equ3.6}
\end{equation}
Equations \ref{equ3.5} and \ref{equ3.6} constitute the compensation unit for the fusion of core features and filtering of irrelevant features.
Features $O_{rgb}$ are fed into the attention generator to generate modal-specific attention scores which are used to calculate modal-level 
saliency consistency loss and represent probability that snippets belong to the foreground. When the main modality is optical flow and the auxiliary 
modality is raw images, the procedure is the same as above, with output features $O_{flow}$. 

Once obtaining complementary enhanced features based on CSCM, achieving modal-level consistency becomes of paramount importance. An intuitive approach may involve directly computing the feature similarity between two modalities while maintaining a high degree of similarity, this direct method
is unsuitable for our work. The issue lies in the sparse distribution of expressions over time, and a direct approximation could result in the loss of crucial features.

Hence, we opt for an indirect approach to generate temporal attention scores using the enhanced features. There are two primary reasons for this choice. 
First, we generate proposals based on higher temporal attention scores during testing, enabling these temporal scores to serve as pseudo-labels for guiding 
subsequent tasks. Second, these temporal attention scores reflect the probability of belonging to a positive sample \cite{islam2021hybrid, hong2021cross}. 
Furthermore, these scores encompass global representation within the temporal dimension. Consequently, we propose utilizing temporal attention scores to 
achieve our ultimate modal-level consistency.

Given that our augmented features originate from salient regions in the two modalities, we define this congruence as modal-level saliency consistency. 
Specifically, we utilize $O_{rgb}$ and $O_{flow}$ to generate modal-specific temporal attention scores. The aim is to capture significant correlations between the two modalities, mitigate suboptimalities arising from their differences, 
and harness the inherent strengths of each modality. Moreover, strong modal coherence implies an attentional score with robust representational power, 
thereby enhancing its utility for downstream tasks.

\noindent \textbf{Video-level Distribution Consistency Strategy.}
Contrary to the general TAL \cite{lin2017single, yang2020revisiting, zhao2017temporal, zhang2020asfd}, expression spotting heavily relies on 
the sample distribution of different classes \cite{qu2017cas, yap2020samm}. We observe that the distribution of MEs in long, untrimmed face 
videos is sparser than that of MaEs on the CAS(ME)$^2$ \cite{qu2017cas} and SAMM-LV \cite{yap2020samm} datasets, because MEs are more challenging 
to evoke than MaEs \cite{li2022cas}. Furthermore, the foreground occupies a smaller portion of the video compared to the background \cite{qu2017cas, 
yap2020samm}, because emotions are frequently expressed in only a few specific (sparse) frames or intervals in a video \cite{xu2016heterogeneous}. 
Therefore, the difference in sample distributions facilitates the distinguishing 
of snippets belonging to the foreground or background in our model.

\begin{figure}[tbp]
  \centering
  \includegraphics[width=\linewidth]{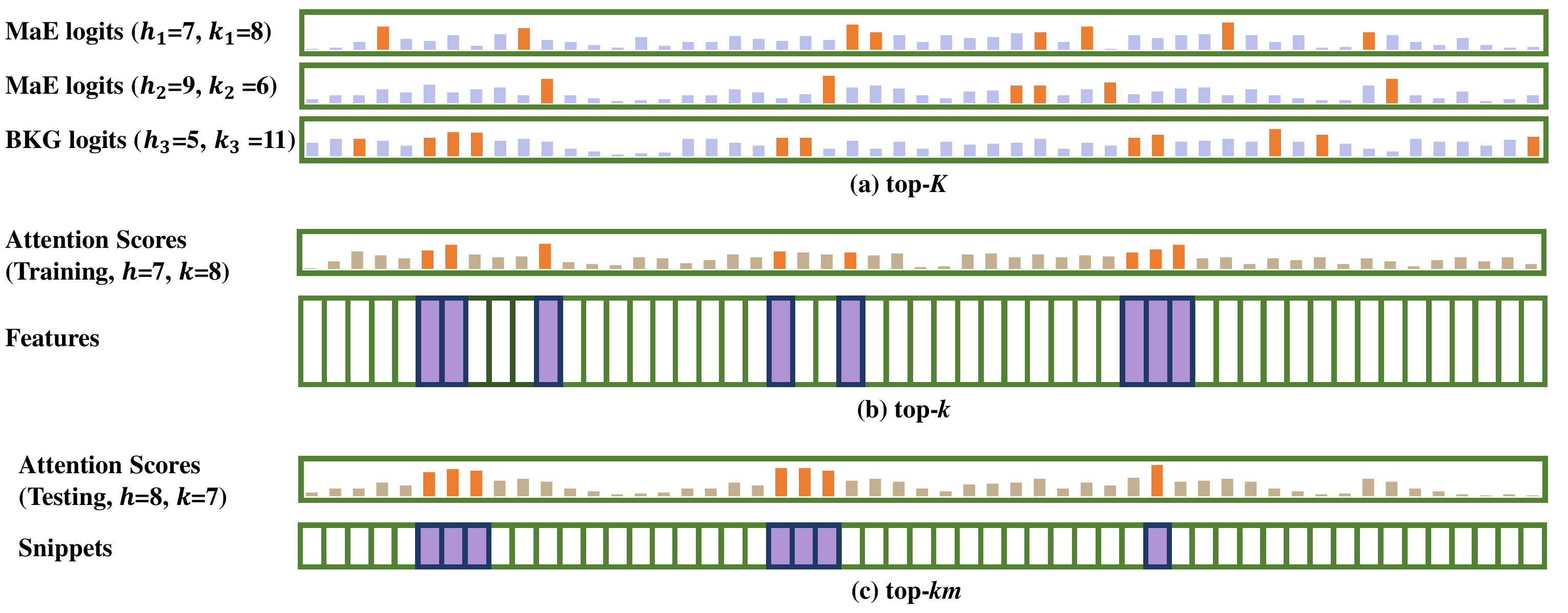}
  \caption{Schematic diagram of the three top-k strategies. We assume the existence of 56 video snippets, each associated with an attention score, as well as the probability-related 
  logits of MEs, MaEs, and backgrounds, respectively. $h$ is used to determine $k$ in Equation \ref{equ3.7}. 
  (a)The top-$K$ strategy is used to select snippet-level logits from each video for calculating class-related 
  probabilities in video-level distribution consistency strategy during training. Here, $K$ contains a number of $k_i$(the number of $i$ equals to the number of categories). 
  (b)The top-k strategy is used to select snippet features corresponding to top-k attention scores in segment-level feature consistency strategy during training. Here, $k$ 
  contains one number which we choose the larger value belonging to the foreground in $K$. (c)The top-$km$ strategy is used to select snippets corresponding to top-$km$ 
  attention scores during testing.}
  \label{fig4}
\end{figure}

\textcolor{black}{Previous WTAL frameworks \cite{hong2021cross, narayan20193c, islam2021hybrid, lee2020background} have generally taken the same values in the top-$K$ \footnote{\textcolor{black}{As shown 
in Figure \ref{fig4}(a)}, the top-$k$ contains a parameter to be specified, while top-$K$ contains $k_1, \ldots, k_n$ where the $i$-th class corresponds to the parameter of $k_i$.}} to 
select the snippet-level temporal logits of T-CAMs along the temporal dimension for different categories and the background class. \textcolor{black}{Here snippet-level temporal logits 
signify the eigenvalues of each snippet pertaining to different classes along the temporal dimension.} These selected 
logits can be used to calculate class-specific average values for computing MIL losses. Compared with the above WTAL approaches with multiple classifications, 
WES is limited to two types, i.e., ME and MaE. Therefore, the distribution of MaEs and MEs can be incorporated into the model training by setting different 
values in the top-$K$ to sample the snippet-level logits of T-CAMs along the class dimension in WES (In the snippet spotting of Figure \ref{fig2}, some branches 
must consider the distribution of the background.) This strategy is defined as video-level distribution consistency in our MIL-based MC-WES framework.

\textcolor{black}{Specifically}, we adopt the top-$K$ temporal average pooling strategy \cite{paul2018w, islam2021hybrid, hong2021cross} and the sampling rate $k_i\in \mathbb{R}^{n+1}$ 
for the $i$-th category is
\begin{equation}
  k_i = \max (1, \lfloor \frac{T}{h_i}\rfloor), 
  \label{equ3.7}
\end{equation}
where $T$ is the number of snippets, and the $h_i$ are the predefined parameters to calculate the sampling rate $k_i$. We denote the snippet-level logits for the 
$j$-th snippet as $s_{ij}$ in T-CAMs $\mathcal{S}$. Our video-level class-wise logits $u_i$ for the $i$-th category is obtained by pooling the snippet-level 
logits corresponding to the top-$k_i$ snippet indexes, 
% \begin{equation}
%   u_i = \underset{|M|=k_i} {\max_{M \subset \{1,\ldots,T \}}}\frac{1}{k_i}  {\textstyle \sum_{j=1}^{k_i}} s_{ij}.
%   \label{equ3.8}
% \end{equation}
\textcolor{black}{\begin{equation}
  u_i =  \frac{1}{k_i} {\sum_{j \in \mathrm{top \text{-}}  k_i \ \mathrm{indexes} }} s_{ij}.
  \label{equ3.8}
\end{equation}}
A softmax function is applied to obtain the video-level class probabilities $p_i = \frac{\exp (u_i)}{\sum_{i = 1}^{n+1} \exp (u_i)} $ 
along the class dimension, which are used to calculate MIL losses.

\noindent \textbf{Label-level Duration Consistency Strategy.}
Many habitual and unconscious actions, such as blinking, pursing the lips, and shaking the head, constitute background intervals, which sometimes 
produce significant inter-frame and interval differences. Foreground intervals also contain such differences due to the persistent change 
in the face. \textcolor{black}{As the intensity and range of facial movements are difficult to quantify \cite{lu2022more}, previous frameworks \cite{yu2021lssnet, wang2021mesnet, 
he2022micro} have chosen the durations of MaEs and MEs as classification benchmarks.} MEs are shown to be challenging to learn due to their short 
duration and low intensity with low confidence \cite{wang2021mesnet, yu2021lssnet}.

\textcolor{black}{Based on existing ME datasets, we observe that the ME intervals cover only a limited number of snippets. For example, as the durations of MEs are 
shorter than 0.5 second \cite{paul2007emotions}, if each snippet contains 8 frames, MEs can cover 2 or 3 snippets when the video is at 30 FPS.
In WTAL frameworks \cite{hong2021cross, islam2021hybrid, lee2020background, lee2021learning, liu2019completeness}, we use \textcolor{black}{attention} scores to measure 
the probability of a snippet belonging to the foreground intervals. Therefore, a group of snippets with a larger average attention score in a 
certain neighborhood range (i.e., the number of snippets covered) refers to a proposal interval. 
Consistent with previous research \cite{islam2021hybrid, hong2021cross}, attention scores are inversely proportional to the probability of a snippet belonging 
to the background. Thus, snippets with lower probability score have a higher probability of being classified as background.} 

\textcolor{black}{Given the above analysis, we take a video containing 15 snippets as example, the first 5 of which belong to MaE, the $10$-th and $11$-th belong to ME, and 
the remaining belong to the background. 
When we select two snippets with a distance of 1 and compute the average of their corresponding attention scores, we can observe that the average score belonging to 
ME (the $10$-th and $11$-th snippets) has a significant difference from both the left and right neighbors. In contrast, the average score belonging to MaE or the background 
tends to be either small at one end and large at the other, or small at both ends. Therefore, we filter these consecutive larger differences to localize the potential ME snippets. 
Furthermore, we classify significant deviations as intra-background and MaE-background differences, and slight deviations as intra-MaE differences.}

\textcolor{black}{To obtain a video sequence that does not contain MEs and then calculate the MIL loss for this video, we readily exclude potential ME intervals by filtering deviations 
from the average attention scores of consecutive snippets.} The remaining snippets are labeled as pseudo MaE and background snippets to refine the model. We define this strategy 
as the label-level duration consistency with pseudo labels.

Specifically, we calculate the mean attention score of the $j$-th snippet in a certain neighborhood range as:
\begin{equation}
  \mathcal{Q} _j = \frac{\sum_{t=j}^{t + \eta} (\mathcal{A}_t)}{\eta },
  \label{equ3.9}
\end{equation}
where $\mathcal{A}_t$ is the snippet-level attention score for the $t$-th snippet, and $\eta$ is the neighborhood range. Then we compute the interval-level 
adjacent deviation $\Delta_j = | \mathcal{Q}_{j+1}-\mathcal{Q}_{j} \vert$ for the $j$-th snippet. Based on deviation filtering, we set up a mask matrix,
\begin{equation}
  \mathcal{M} _j =
  \begin{cases}
    0, & if  \quad \omega_l\overline{\mathcal{Q}} < \Delta_j  < \omega_u\overline{\mathcal{Q}}\\
    1, & otherwise, 
    \end{cases}
  \label{equ3.10}
\end{equation}
where $\overline{\mathcal{Q}}$ is the mean deviation, $\omega_l$, and $\omega_u$ are the parameters defining the respective lower and upper bounds. \textcolor{black}{When the 
adjacent deviation $\Delta_j$ for the $j$-th interval is less than $\omega_l\overline{\mathcal{Q}}$, we can localize the background and MaE-background snippets, 
and when it is greater than $\omega_u\overline{\mathcal{Q}}$, we can localize the intra-MaE snippets.}

By integrating the mask matrix $\mathcal{M}$ and class-specific T-CAMs $\mathcal{S}$, the logits $\widetilde{\mathcal{S}} = \mathcal{M} \odot \mathcal{S}$ 
corresponding to the background and MaE snippets are obtained. This strategy involves creating pseudo-labels using attention scores, supervising subsequent steps using these labels.

\noindent \textbf{Segment-level Feature Consistency Strategy.}
As our MC-WES framework is only given with video-level supervised information, modeling the correlation between videos of the same category becomes 
particularly significant. Previous approaches \cite{paul2018w, hong2021cross} only use classification-related features to compare video pairs with partially 
consistent labels to reduce intra-class variation and increase inter-class variation. \textcolor{black}{This is carried out by activating T-CAMs $\mathcal{S} \in \mathbb{R} ^{T\times (n+1)}$ 
along the temporal dimension (i.e., $T$) with a softmax function to generate activity portions in a co-activity similarity loss \cite{paul2018w, hong2021cross}.
Then, these generated portions are merged with corresponding video features in the video pair to calculating similarities. However, once the number $T$ 
of snippets is too larger, the activity portions by normalizing the class-specific logits may be overly smooth.} Moreover, the diversity of contexts can lead 
to intra-class variation, which can interfere with the computation of inter-video correlations if all features of a video pair are used.

Therefore, we utilize the \textcolor{black}{top-$k$} (\textcolor{black}{as shown in Figure \ref{fig4}(b), here we use the sampling rate $k$ of MaEs from Equation \ref{equ3.7}}) localization-related attention scores to select 
corresponding classification-related features and logits to implement the segment-level feature consistency strategy in a video pair (e.g., $v_1$ and $v_2$).
This operation is used to localize potential ME and MaE snippets \cite{hong2021cross, paul2018w}.
Then we compute the snippet-level similarity of the selected features of $v_1$ and the other features of $v_2$. Each snippet-level maximum similarity 
of $v_2$ is required to match as closely as possible with the corresponding snippet-level attention score of $v_2$. 

\textcolor{black}{Particularly}, we \textcolor{black}{select} some videos from each batch to construct \textcolor{black}{$N_p$ video pairs $(v_{d1}, v_{d2}), d=1,\ldots, N_p$}, where the labels in the two videos of 
each pair are at least partially the same. \textcolor{black}{Then, we treat the indexes corresponding to top-$k$ attention scores $\mathcal{A}_{d1}$ of video $v_{d1}$ as guiding 
labels to help select the logits $u_{d1}' \in \mathbb{R}^{k \times n}$ and the fused features $f_{d1}' \in \mathbb{R}^{k \times D}$ from video $v_{d1}$.} Note that the 
selected logits do not contain those associated with background classes. The selected logits $u_{d1}'$ are activated separately by a softmax function along the 
class dimension (i.e., $n$), and integrated with the selected features $f_{d1}'$ to obtain category-level features $fs_{d1} \in \mathbb{R}^{k\times D \times n}$. 
The similarity $h_{d2} \in \mathbb{R}^{k \times T \times n}$ of video $v_{d2}$ is calculated as
\begin{equation}
  h_{d2} = \varepsilon ((f_{d2} \otimes (fs_{d1})^T)^T) 
  \label{equ3.11}
\end{equation}
where ``$\otimes$'' denotes matrix multiplication, and $f_{d2} \in \mathbb{R}^{T\times D}$ is the feature of video $v_{d2}$, $\varepsilon$ is used to 
modulate the results of matrix multiplication along the top-$k$ dimension. We select the maximum similarity $ms_{d2}$ along the top-$k$ dimension,
\begin{equation}
  ms_{d2} = \max(h_{d2k}). 
  \label{equ3.12}
\end{equation}
The maximum similarities $ms_{d2} \in \mathbb{R}^{ T \times n}$ and attention scores $\mathcal{A}_{d2} \in \mathbb{R}^{T}$ of video $v_{d2}$ are optimized 
to match as closely as possible based on the matching function $cs$, defined as
\begin{equation}
  cs_{d2} =  \sum_{i=1}^{n} \sum_{j=1}^{T} \frac{{y_{di}\cdot \mathcal{A}_{d2j}}^T \otimes ms_{d2ji}}{\sum_{j = 1}^{T} \mathcal{A}_{d2j}}, 
  \label{equ3.13}
\end{equation}
where $y_d = y_{d1} \odot y_{d2}$ consists of the labels of the videos of $vd_1$ and $vd_2$ without the background class. This similarity function is analogous 
to cosine similarity. When the roles of the two videos are exchanged, we can use the same workflow to obtain $cs_{d1}$.

\subsection{Model Training}
\noindent \textbf{\textcolor{black}{Modal-level Saliency Consistency Loss.}}
To maintain consistency of information across modalities and strengthen the effectiveness of CSCM, we follow CO$_2$-Net \cite{hong2021cross} to apply mutual 
learning loss on two modal-specific attention scores with the mean square error (MSE) function,
\textcolor{black}{\begin{equation}
  \mathcal{L}_{sc} =  \frac{1}{2T} (\| \mathcal{A}_{rgb} -\rho (\mathcal{A}_{flow})\Vert_2 ^2 + \| \rho(\mathcal{A}_{rgb}) - \mathcal{A}_{flow}\Vert_2^2),
  \label{equ3.14}
\end{equation}}
where $\rho(\cdot) $ is the gradient detachment function and $\|\cdot\Vert_2$ is the L2-norm function.

\noindent \textbf{\textcolor{black}{Attention-guided Distribution Consistency Loss.}}
As shown in Figure \ref{fig2}, there are three branches in snippet spotting to process the logits of T-CAMs $\mathcal{S}$, of which the first two relate only 
to the video-level distribution consistency strategy. Therefore, we calculate the MIL loss in the first branch by utilizing the original logits and the video-level 
distribution consistency strategy to produce video-level class probability scores $p_i^1$ with the top-$k_i$ in Equation \ref{equ3.7} for the $i$-th category,
\textcolor{black}{\begin{equation}
  \mathcal{L}_{dc}^1 =  -\sum_{i=1}^{n+1} y_i^1 \log (p_i^1),
  \label{equ3.15}
\end{equation}}
where $y_i^1$ consists of the ground truth labels with the background class, i.e., $y_{n+1}^1 = 1$.

The middle branch applies the attention score $\mathcal{A}$ to inhibit the background snippets in T-CAMs $\mathcal{S}$. Then we use the video-level distribution 
consistency strategy to generate video-level class probability scores $p_i^2$ with processed logits of T-CAMs $\widehat{\mathcal{S}}$. Therefore, the loss function 
of the middle branch is
\textcolor{black}{\begin{equation}
  \mathcal{L}_{dc}^2 =  -\sum_{i=1}^{n+1} y_i^2 \log (p_i^2),
  \label{equ3.16}
\end{equation}}
where $y_i^2$ is the ground truth label without the background class, i.e., $y_{n+1}^2 = 0$.

Furthermore, due to the observation that the positive samples are sparsely distributed in a long video \cite{nguyen2018weakly, xu2016heterogeneous}, we utilize the 
L1-norm to guarantee sparsity of positive samples,
\textcolor{black}{\textcolor{black}{\begin{equation}
  \mathcal{L}_{sl} =  \frac{1}{3T} (\| \mathcal{A} \Vert_1  + \| \mathcal{A}_{rgb} \Vert_1 + \| \mathcal{A}_{flow} \Vert_1).
  \label{equ3.17}
\end{equation}}}

\noindent \textbf{\textcolor{black}{Attention-guided Duration Consistency Loss.}}
As shown in Figure \ref{fig2}, the third branch integrates the mask matrix $\mathcal{M}$ generated by the label-level duration consistency strategy and the original 
T-CAMs $\mathcal{S}$ to produce video-level class probability scores $p_i^3$ with the video-level distribution consistency strategy. Therefore, the loss of this branch 
is defined as:
\textcolor{black}{\begin{equation}
  \mathcal{L}_{dc}^3 = -\sum_{i=1}^{n+1} y_i^3 \log (p_i^3),
  \label{equ3.18}
\end{equation}}
where $y_i^3$ represents the ground truth labels with the background class and without ME class.

\noindent \textbf{\textcolor{black}{Attention-guided Feature Consistency Loss.}}
To merge localization and classification information, the segment-level feature consistency strategy is designed to calculate the similarity $cs_{d1}$ and $cs_{d2}$ for 
the $d$-th video pair. Accordingly, we assume that for a video pair, video-level labels are present in both videos as valid labels (without the background class). 
Then \textcolor{black}{we count $N_s$, the number of valid labels across all video pairs in one batch. For example, if video $v_{d1}$ contains MaEs and MEs ($y_{mae}=1, y_{me}=1$), and 
video $v_{d2}$ contains only MaEs ($y_{mae}=1, y_{me}=0$), the labels of the two videos are partially same, having one valid label.} Therefore, the attention-guided 
feature consistency loss is calculated as:
\begin{equation}
  \mathcal{L}_{fc} =  1 - \frac{1}{2N_s} \sum_{d = 1}^{N_p}  (cs_{d1} + cs_{d2}).
  \label{equ3.19}
\end{equation}
where $N_p$ is the total number of the video pairs in one batch.

\noindent \textbf{Final Joint Loss.}
Following previous works \cite{islam2021hybrid, hong2021cross}, our MC-WES framework uses the guide loss function to make the attention score inverse to the probability 
that a snippet belongs to the background. In this end, we first calculate the probability of a snippet being in foreground intervals,
\begin{equation}
  p_f = 1- p_{(n+1)},
  \label{equ3.20}
\end{equation}
where $p_{(n+1)} $ is the probability of a snippet belonging to the background class. Then the guide loss function is
\begin{equation}
  \mathcal{L}_{gl} = \frac{1}{3T} ( \| p_f - \mathcal{A}\Vert_1  + \| p_f - \mathcal{A}_{rgb}\Vert_1  + \| p_f - \mathcal{A}_{flow}\Vert_1),
  \label{equ3.21}
\end{equation}
where $\|\cdot\Vert_1$ is the L1-norm function, $\mathcal{A}_{rgb}$ and $\mathcal{A}_{flow}$ are respectively the attention scores of the 
snippets of raw images and optical flow, and $\mathcal{A}$ is the snippet-level mean of $\mathcal{A}_{rgb}$ and $\mathcal{A}_{flow}$.

Finally, we combine the above loss functions to form the final optimization function for the whole framework,
\textcolor{black}{\begin{equation}
  \begin{split}
    \mathcal{L} & =  \mathcal{L}_{sc} + \mathcal{L}_{dc}^1 + \mathcal{L}_{dc}^2 + \lambda_1\mathcal{L}_{dc}^3  \\
    & + \lambda_2\mathcal{L}_{fc} + \lambda_3\mathcal{L}_{sl} + \lambda_4\mathcal{L}_{gl} ,
  \end{split}
  \label{equ3.22}
\end{equation}}
where $\lambda_1$, $\lambda_2$, $\lambda_3$, and $\lambda_4$ are predefined hyperparameters.

\subsection{Expression Spotting}
\label{Section 3.5}
During testing, previous works \cite{islam2021hybrid, hong2021cross} use a multi-threshold method to spot the final proposals. Specifically,  
upper and lower bounds, $\tau_u$ and $\tau_l$, are first set, along with level $N_l$ to 
produce a one-dimensional threshold vector, uniformly divided between $\tau_u$ and $\tau_l$. Then a threshold is selected from this threshold vector to 
filter snippets whose class-agnostic attention scores are larger than the threshold. These filtered snippets are sorted based on timestamps. As a result, these 
snippets corresponding to the consecutive timestamps are the final proposals 

Instead, we use a series of consecutive integers to build a set $M'$, each value of which is utilized to obtain the number $km$ (as shown 
in Figure \ref{fig4}(c)) in Equation \ref{equ3.7}. We then select snippets whose class-agnostic attention scores belong to top-$km$ class-agnostic attention scores. These 
filtered snippets are then used to generate proposals with the similar steps to the above multi-threshold method. 
We define this method as multi-top. 
The performances of our multi-top method and the existing multi-threshold method will be further compared and discussed in Section \ref{Section 4.4.5}.

Following AutoLoc \cite{shou2018autoloc}, the generated proposals are defined as $[f_{on}, f_{off}]$, with varying durations. To spot short-duration intervals and generate as 
few proposals as possible, we set 15 consecutive integers in $M'$ set. Then, we define the duration of the $i$-th proposal as $dp_i = f_{on}^i - f_{off}^i + 1$, 
and calculate the class-specific score $\phi_{ij}$ for the $j$-th category with the suppressed T-CAMs $\widehat{\mathcal{S}}$,
\begin{equation}
  \phi_{ij} =  \phi_{ij}^{inner} -\phi_{ij}^{outer} + \varsigma p_{ij}.
  \label{equ3.23}
\end{equation}
\begin{equation}
  \phi_{ij}^{inner}=  \frac{1}{dp_i} {\sum_{t = f_{on}^i}^{f_{off}^i} s_{ijt}},
  \label{equ3.24}
\end{equation}
\begin{equation}
  \phi_{ij}^{outer} =  \frac{1}{2\psi dp_i} ({\sum_{t = f_{on}^i-\psi dp_i }^{f_{on}^i} s_{ijt}}  +  {\sum_{t = f_{off}^i}^{f_{off}^i + \psi dp_i} s_{ijt}}),
  \label{equ3.25}
\end{equation}
where $\varsigma $ is a hyperparameter related to the logits of all proposals, $p_{ij}$ is the video-level class logit for the $j$-th category, $s_{ijt}$ is the 
snippet-level class logit for the $j$-th category, $\psi$ is a hyperparameter related to the durations of all proposals, $\phi_{ij}^{inner}$ is the inner 
class logit for the $j$-th category, which is the mean logit from timestamp $f_{on}^i$ and $f_{off}^i$, and $\phi_{ij}^{outer}$ is the outer class logit for 
the $j$-th category, which is from the mean of the corresponding logits after expanding $\psi dp_i$ timestamps since $f_{on}^i$ towards the video beginning and $f_{off}^i$ 
towards the video end, respectively. The essence of the class-specific logit $\phi_{ij}$ for the $j$-th category is the outer-inner score of AutoLoc \cite{shou2018autoloc}. 

\textcolor{black}{Because there are two types of expressions (i.e., ME and MaE), each proposal generated by the attention scores has two class-specific scores. Thus, we process 
$2n_e$ generated expression proposals $(f_{on}^{ij}, f_{off}^{ij}, y_{ij}, \phi_{ij})$, where $i=1,\ldots,n_e$, $j=1,2$ and $n_e$ is the number of the generated proposals 
for each class, using non-maximum suppression (NMS) \cite{neubeck2006efficient} to eliminate redundant proposals.}

\section{Experiments}

\subsection{Datasets}
\label{Section 4.1}
We evaluate our MC-WES framework on three popular spotting datasets: CAS(ME)$^2$ \cite{qu2017cas}, SAMM-LV \cite{yap2020samm}, and CAS(ME)$^3$ dataset \cite{li2022cas}. 
CAS(ME)$^2$ dataset consists of 98 long videos at 30 FPS, each with an average of 2940 frames and 96$\%$ background, annotated with 57 MEs and 300 MaEs from 22 subjects. 
SAMM-LV dataset consists of 224 long videos at 200 FPS, each with an average of 7000 frames and 68$\%$ background, annotated with 159 MEs and 340 MaEs from 32 subjects. 
Furthermore, CAS(ME)$^3$ dataset comprises 956 videos at 30 FPS, each with an average of 2600 frames and 84$\%$ background, encompassing 207 instances of MEs 
and 2071 instances of MaEs. \textcolor{black}{It's worth noting that there are differences in labeling principles between these three datasets. For instance, the CAS(ME)$^2$ and CAS(ME)$^3$ 
datasets do not contain MaE ground truth intervals that exceed 4.0 seconds, whereas some ground truth intervals in the SAMM-LV dataset extend beyond 20.0 seconds, deviating 
far from Ekman's observation that normal MaEs typically last within 4.0 seconds \cite{ekman2003darwin}. Specifically, these abnormally long MaE ground truths in a video may 
cause some short ME ground truths in this video to be neglected. In essence, these video-level labels for weak supervision are somewhat noisy.}

\subsection{Evaluation Metrics}
\label{Section 4.2}
Following Micro-Expression Grand Challenge (MEGC) 2019 \cite{see2019megc} and 2022 \cite{li2022megc2022}, we employ the intersection over union (IoU) method to select 
eligible expression proposals that are defined as true positive (TP) samples. The IoU between a spotted proposal $E_s$ and a ground truth interval $E_{gt}$ is calculated as:
\begin{equation} 
  \frac{E_s\bigcap E_{gt}}{E_s\bigcup  E_{gt}} \geq k_{eval},
  \label{equ3.26}
\end{equation} 
where $k_{eval}$ is the evaluation threshold, which is commonly set to 0.5 for expression spotting \cite{yu2021lssnet}. Hence, \textcolor{black}{when a proposal matches a ground truth interval 
with an IoU greater than or equal to 0.5, we classify it as a TP sample. Any proposals not meeting this criterion are categorized as False Positive (FP) samples. After counting 
the numbers of TP and FP samples, we can compute the overall precision, overall recall and overall F1-score based on the evaluation metrics used in MEGC 2020 \cite{see2019megc} and 2022 \cite{li2022megc2022}.}

\textcolor{black}{Furthermore, existing methods \cite{pan2020local, liong2022mtsn, wang2021mesnet} adopt a testing criterion where proposals lasting longer than 0.5 second are classified as ME proposals, 
while the remaining are labeled as MaE proposals. Considering CAS(ME)$^2$ and SAMM-LV have the frame rates of 30 and 200 FPS, respectively, a duration of 0.5 second corresponds precisely 
to 15 frames and 100 frames for these two datasets, respectively. Some proposals with durations up to 1.0 second also match the ground truths of MEs and will be classified as MEs, when calculating the overall optimal F1-score. Moreover, a duration of 1.0 second corresponds precisely to 30 frames and 200 frames for these two datasets, respectively.}  
To evaluate the impact of these proposals with durations ranging from 0.5 to 1.0 second on ME spotting, \textcolor{black}{F1-ME (0.5) and F1-ME (1.0) are defined respectively as the F1-scores associated with MEs 
for the proposals with durations below 0.5 and 1.0 second. That is, we select all the proposals with duration below 0.5 second to compute F1-ME (0.5), and select all the proposals 
with duration below 1.0 second to compute F1-ME (1.0).} Following the previous work \cite{he2020spotting, he2022micro}, \textcolor{black}{when we get the set of proposals corresponding to the overall 
optimal F1-score, we can figure out the proposals with durations below 0.5 second as ME proposals from this set to calculate the F1-score specific to MEs, i.e., F1-ME (p).}

In summary, we can calculate three F1-scores for MEs, defined as F1-ME(0.5), F1-ME(1.0), and F1-ME(p). F1-ME(p) and F1-ME(0.5) are employed to assess whether all ME 
TP samples in the overall proposal set are present in the optimal proposal set. If F1-ME(p) and F1-ME(0.5) are close, it indicates that the model can spot not only the majority of TP 
samples but also the majority of ME TP samples. Furthermore, \textcolor{black}{if F1-ME (0.5) is much larger than F1-ME (1.0), it indicates that proposals with durations between 0.5 and 1.0 second are 
very few and has little effect on ME spotting.}

\subsection{Implementation Details}
On the CAS(ME)$^2$ dataset, we sample every continuous non-overlapping 8 frames as a snippet to split each video and optical flow, and on the SAMM-LV dataset,
we take a fix duration of 32 frames as a snippet. Then we apply the I3D model to extract 1024-dimension features for each snippet. \textcolor{black}{During training, 
for the purpose of facilitating the training process, we randomly sample 250, 300 and 380 snippets for each video of the CAS(ME)$^2$, CAS(ME)$^3$ and SAMM-LV dataset, respectively.} 
During testing, we take all snippets for each video. 

We use Adam \cite{kingma2014adam} as the optimizer, and train the model with 1000 iterations for each dataset. The batch size is set to 10. \textcolor{black}{In each batch, six videos (i.e., $N_p$=6) 
are used to construct three (i.e., $d$=3) video pairs to implement the segment-level feature consistency strategy.} For CAS(ME)$^2$, we set $\lambda_1 = \lambda_2 = 0.5$, $\lambda_3 
= \lambda_4 = 0.8$, a learning rate of 0.0005 during training, and $\varsigma = 0.15$ and $\psi = 0.25$ during testing. For SAMM-LV, we set $\lambda_1 = \lambda_2 = 0.5$, 
$\lambda_3 = \lambda_4 = 0.7$, and a learning rate of 0.0008 during training, and $\varsigma = 0.5$ and $\psi = 0.25$ during testing. 
\textcolor{black}{Regarding the settings for selecting potential ME snippets in label-level duration consistency strategy, it's important to note that these settings are based on our empirical tuning. 
Specifically, we set the larger $\omega_l$ and $\omega_u$ for training on SAMM-LV, and smaller $\omega_l$ and $\omega_u$ on CAS(ME)$^2$, mainly because the images in SAMM-LV are grayscale
maps, the empirical differences between neighboring snippets are relatively weaker compared with that of CAS(ME)$^2$ containing images all in RGB format. Particularly, to enlarge these fine-grained differences 
and capture potential ME snippets, we set $\omega_l = 1.2$ and $\omega_u=1.4$ for CAS(ME)$^2$, and $\omega_l = 1.5$ and$\omega_u=1.8$ for SAMM-LV.}
To remove redundant proposals, we use NMS \cite{neubeck2006efficient} on both datasets, with a threshold of 0.01. \textcolor{black}{When calculating precision and recall rates, we determine 
the categories of proposals based on their temporal durations \cite{see2019megc}. Additionally, we employ a truncation threshold, set to be 0.1,
to filter out proposals with a confidence score below this threshold.}

\subsection{Ablation Study}
We utilize the leave-one-subject-out (LOSO) learning strategy to train our model and generate proposals. After training, we use NMS to remove redundant proposals. 
The remaining proposals are used to calculate the recall rate, precision rate, and F1-score on the CAS(ME)$^2$, CAS(ME)$^3$, and SAMM-LV datasets. 

\subsubsection{Effect of Modal-level Consistency}
To alleviate the modal-level information discrepancy from the raw image and optical flow, we train our model on the CAS(ME)$^2$ and SAMM-LV datasets using
different multi-modal feature fusion methods, including direct concatenating, CCM \cite{hong2021cross}, and our proposed CSCM with the same configuration.

The results of Table \ref{tab1} and \ref{tab2} show that CSCM can significantly improve the recall rate, precision rate, and F1-score 
of the model on both datasets compared with previous methods such as direct concatenating and CCM. In particular, compared with direct concatenating, 
CCM improves ME and MaE spotting, but the improvement by CSCM is more significant. These results suggest that a network derived from video action 
recognition, e.g., I3D, may produce information redundancy in the extracted features that affects the learning of the spotting model.
Moreover, inter-modal consensus modules like CCM can help eliminate redundant information and improve results, but there is still inter-modal feature 
inconsistency, which may degrade the model's performance. In contrast, our proposed CSCM effectively removes information redundancy and enables modal-level 
consistency to alleviate inter-modal gaps.

On both datasets, our proposed CSCM leads to a noticeable increase in the spotting results of MEs over CCM and direct concatenation, which cannot make the 
proposals with the optimal F1-score contain the most MEs, with the result that F1-ME(p) is not close to F1-ME(0.5) or much larger than F1-ME(1.0). With CSCM, 
the best result of F1-ME(0.5) is equal to F1-ME(p) and much larger than F1-ME(1.0) on the two datasets, indicating that it can contribute meaningfully to 
capturing the most MEs in the proposals.

\begin{table}[htbp]
  \centering
  \small 
  \setlength\tabcolsep{10pt}
  \caption{Performances with different multi-modal feature fusion methods on the CAS(ME)$^2$ dataset.}
    \begin{tabular}{c|cccccc}
    \toprule
    \multirow{2}[4]{*}{Measure} & \multicolumn{3}{c}{Fusion Method} \\
    \cmidrule{2-4}          & Concatenate & CCM & CSCM  \\  
    \midrule
    F1-ME(0.5)  &0.118 &0.141 &\textbf{0.167}  \\
    F1-ME(1.0)  &0.091 &0.092 &\textbf{0.108}  \\
    F1-ME(p)    &0.034 &0.114 &\textbf{0.169}  \\                                                                                                                          
    Recall      &0.143 &0.202 &\textbf{0.266}  \\
    Precision   &0.378 &0.283 &\textbf{0.415}  \\
    F1-score    &0.207 &0.236 &\textbf{0.324}  \\
    \bottomrule
    \end{tabular}
  \label{tab1}
\end{table}

\begin{table}[htbp]
  \centering
  \small 
  \setlength\tabcolsep{10pt}
  \caption{Performances with different multi-modal feature fusion methods on the SAMM-LV dataset.}
    \begin{tabular}{c|cccccc}
    \toprule
    \multirow{2}[4]{*}{Measure} & \multicolumn{3}{c}{Fusion Method} \\
    \cmidrule{2-4}          & Concatenate & CCM & CSCM  \\
    \midrule
    F1-ME(0.5)  &0.110 &0.097 &\textbf{0.135}  \\
    F1-ME(1.0)  &0.048 &0.040 &\textbf{0.055}  \\
    F1-ME(p)    &0.093 &0.048 &\textbf{0.135}  \\
    Recall      &0.218 &0.216 &\textbf{0.263}  \\
    Precision   &0.136 &0.172 &\textbf{0.178}  \\
    F1-score    &0.168 &0.192 &\textbf{0.212}  \\
    \bottomrule
    \end{tabular}
  \label{tab2}
\end{table}

\begin{table*}[htbp]
  \centering
  \small 
  \setlength\tabcolsep{1.5pt}
  \caption{Performances with different top\_$K$ temporal average pooling strategy on the CAS(ME)$^2$ and SAMM-LV datasets. 
    ``$h$'' is the predefined parameters in Equation \ref{equ3.7}.}
    \begin{tabular}{c|cccccc|cccccc}
    \toprule
    \multirow{2}[4]{*}{$h$} & \multicolumn{6}{c|}{CAS(ME)$^2$}                        & \multicolumn{6}{c}{SAMM-LV} \\
\cmidrule{2-13}          &F1-ME(0.5) &F1-ME(1.0) &F1-ME(p) &Recall &Precision &F1-score &F1-ME(0.5) &F1-ME(1.0) &F1-ME(p) &Recall &Precision &F1-score \\
    \midrule
    $[5,5,5]$ &0.034 &0.034 &0.027 &0.109 &0.361 &0.168 &0.107 &0.043 &0.074 &0.234 &0.159 &0.189 \\
    $[7,7,7]$ &0.066 &0.057 &0.029 &0.118 &0.400 &0.182 &0.110 &0.043 &0.055 &0.238 &0.162 &0.193 \\
    $[9,9,9]$ &0.034 &0.034 &0.028 &0.112 &\textbf{0.444} &0.179 &0.098 &0.046 &0.066 &0.242 &\textbf{0.178} &0.205 \\
    $[5,7,5]$ &0.116 &0.098 &0.000 &0.148 &0.387 &0.215 &0.104 &0.043 &0.000 &0.176 &0.209 &0.191 \\
    $[7,7,5]$ &0.149 &0.095 &0.000 &0.193 &0.309 &0.238 &0.105 &0.046 &0.049 &0.198 &0.190 &0.194 \\
    $[5,9,5]$ &0.138 &\textbf{0.111} &0.033 &0.179 &0.379 &0.243 &0.111 &0.044 &0.000 &0.220 &0.180 &0.198  \\
    $[7,9,5]$ &\textbf{0.167} &0.108 &\textbf{0.169} &\textbf{0.266} &0.415 &\textbf{0.324} &\textbf{0.135} &\textbf{0.055} &\textbf{0.135} &\textbf{0.263} &\textbf{0.178} &\textbf{0.212} \\
    \bottomrule
    \end{tabular}
  \label{tab3}
\end{table*}

\begin{table}[htbp]
  \centering
  \small 
  \setlength\tabcolsep{5pt}
  \caption{Performances with different durations of snippets on the CAS(ME)$^2$ and SAMM-LV datasets.}
    \begin{tabular}{c|ccc|ccc}
    \toprule
    \multirow{3}[6]{*}{Measure} & \multicolumn{6}{c}{Duration} \\
    \cmidrule{2-7}          & \multicolumn{3}{c|}{CAS(ME)$^2$} & \multicolumn{3}{c}{SAMM-LV} \\
\cmidrule{2-7}          & 4 & 8 & 16 & 16 & 32 & 48 \\
    \midrule
    F1-ME(0.5)  &0.110 &\textbf{0.167} &0.0   &\textbf{0.174} &0.135 &0.105 \\
    F1-ME(1.0)  &0.080 &\textbf{0.108} &0.0   &0.056 &0.055 &\textbf{0.057} \\
    F1-ME(p)    &0.083 &\textbf{0.169} &0.0   &0.118 &\textbf{0.135} &0.088 \\
    Recall      &0.152 &\textbf{0.266} &0.129 &0.194 &\textbf{0.263} &0.212 \\
    Precision   &0.346 &\textbf{0.415} &0.189 &0.134 &\textbf{0.178} &0.162 \\
    F1-score    &0.211 &\textbf{0.324} &0.153 &0.159 &\textbf{0.212} &0.184 \\
    \bottomrule  
    \end{tabular}
  \label{tab4}
\end{table}

\begin{table*}[htbp]
  \centering
  \small 
  \setlength\tabcolsep{5pt}
  \caption{Ablation studies of our model for MaE and ME spotting on the CAS(ME)$^2$ dataset with different combinations of loss.}
    \begin{tabular}{c|ccccccc|cccccc}
    \toprule
    EXP&$\mathcal{L}_{sc}$&$\mathcal{L}_{dc}^1$&$\mathcal{L}_{dc}^2$&$\mathcal{L}_{dc}^3$&$\mathcal{L}_{fc}$&$\mathcal{L}_{sl}$&$\mathcal{L}_{gl}$&F1-ME(0.5)&F1-ME(1.0)&F1-ME(p)&Recall&Precision&F1-score \\
    \midrule
    1     &       &       &       &       &       &$\surd$&       &0.034 &0.032 &0.000 &0.048 &0.050 &0.049 \\
    2     &       &       &       &       &       &       &$\surd$&0.103 &0.068 &0.000 &0.087 &0.077 &0.082  \\
    3     &       &       &       &       &       &$\surd$&$\surd$&0.015 &0.008 &0.000 &0.053 &0.056 &0.054  \\ 
    4     &$\surd$&       &       &       &       &       &       &0.034 &0.027 &0.000 &0.064 &0.071 &0.067  \\
    5     &$\surd$&       &       &       &       &$\surd$&       &0.133 &0.034 &0.000 &0.070 &0.071 &0.070  \\
    6     &$\surd$&       &       &       &       &       &$\surd$&0.015 &0.009 &0.000 &0.053 &0.055 &0.054  \\
    7     &$\surd$&       &       &       &       &$\surd$&$\surd$&0.034 &0.010 &0.000 &0.073 &0.069 &0.071  \\
    8     &$\surd$&$\surd$&       &       &       &$\surd$&$\surd$&0.069 &0.045 &0.000 &0.090 &0.102 &0.095  \\
    9     &$\surd$&       &$\surd$&       &       &$\surd$&$\surd$&0.149 &0.043 &0.034 &0.081 &0.236 &0.121  \\
    10    &$\surd$&$\surd$&$\surd$&       &       &$\surd$&$\surd$&0.113 &0.090 &0.034 &0.199 &0.183 &0.191  \\
    11    &$\surd$&$\surd$&       &$\surd$&       &$\surd$&$\surd$&0.100 &0.069 &0.000 &0.067 &0.202 &0.101  \\
    12    &$\surd$&$\surd$&$\surd$&$\surd$&       &$\surd$&$\surd$&0.091 &0.063 &0.036 &0.157 &0.220 &0.183  \\
    13    &$\surd$&$\surd$&       &       &$\surd$&$\surd$&$\surd$&0.069 &0.027 &0.000 &0.053 &0.186 &0.083  \\
    14    &$\surd$&$\surd$&$\surd$&       &$\surd$&$\surd$&$\surd$&0.121 &0.076 &0.057 &0.154 &0.342 &0.212  \\
    15    &$\surd$&$\surd$&       &$\surd$&$\surd$&$\surd$&$\surd$&0.069 &0.042 &0.0   &0.076 &0.095 &0.084  \\
    16    &$\surd$&$\surd$&$\surd$&$\surd$&$\surd$&$\surd$&$\surd$&\textbf{0.167} &\textbf{0.108} &\textbf{0.169} &\textbf{0.266} &\textbf{0.415} &\textbf{0.324} \\
    \bottomrule  
  \end{tabular}
  \label{tab5}
\end{table*}

\begin{table}[htbp]
  \centering
  \small 
  \setlength\tabcolsep{5pt}
  \caption{Performances with different post-processing on the CAS(ME)$^2$ and SAMM-LV datasets.}
    \begin{tabular}{c|cc|cc}
    \toprule
    \multirow{3}[6]{*}{Measure} & \multicolumn{4}{c}{Post-processing} \\
    \cmidrule{2-5}          & \multicolumn{2}{c|}{CAS(ME)$^2$} & \multicolumn{2}{c}{SAMM-LV} \\
    \cmidrule{2-5}          & top & threshold & top & threshold \\
    \midrule
    F1-ME(0.5)  &\textbf{0.167} &0.078 &\textbf{0.135} &0.092  \\
    F1-ME(1.0)  &\textbf{0.108} &0.069 &\textbf{0.057} &0.050  \\
    F1-ME(p)    &\textbf{0.169} &0.028 &\textbf{0.135} &0.078  \\
    Recall      &\textbf{0.266} &0.162 &\textbf{0.263} &0.206  \\
    Precision   &\textbf{0.415} &0.152 &\textbf{0.178} &0.150  \\
    F1-score    &\textbf{0.324} &0.157 &\textbf{0.212} &0.173  \\
    \bottomrule  
    \end{tabular}
  \label{tab7}
\end{table}

\subsubsection{Hyperparameters in Video-level Consistency}
To merge the prior about the distribution information of different categories, we adopt a video-level distribution consistency strategy with different values
to achieve the top-$K$ temporal average pooling strategy in our MIL-based framework. This strategy is used to calculate the average logits of different classes 
and then obtain the final classification loss. Therefore, we set various combinations of values in $h$ to calculate the sampling rate for different categories 
using Equation \ref{equ3.7}. For example, with a combination $[5,5,5]$ in $h$, the first two values are used to calculate the sampling rate of the MaE and ME logits, 
respectively, and the last to calculate the sampling rate of the background logits.

The results of Table \ref{tab3} show that computing the average logits with the same values (i.e., $[5,5,5], [7,7,7], [9,9,9]$) for different categories 
and the background class in the MIL-based framework reduces the spotting capability of the model, whereas our strategy with different values in $h$ (i.e.,
$[7,9,5]$) achieves better results. Xu et al. \cite{xu2016heterogeneous} points out that most of the frames belong to the background, which is also 
observed in Section \ref{Section 4.1}. Therefore, the background parameter in $h$ must be set to the minimum to ensure that most of the background logits 
are retained for the calculation of the average logits. Moreover, since MEs are distributed more sparsely than MaEs \cite{qu2017cas, yap2020samm}, the 
value of MaE is lower than that of ME in $h$. Using the same part of the values of $h$ (i.e., $[5,7,5], [7,7,5], [5,9,5]$) does not significantly improve 
the results of our spotting model, which shows that our video-level distribution consistency strategy of setting different values based on the 
distribution (i.e., $[7,9,5]$) is valid. 

In addition, our strategy, which utilizes different values in $h$ (i.e., $[7,9,5]$), is evidently more effective in ME spotting. Other configurations 
of values in $h$ (such as $[5,5,5], [7,7,7], [9,9,9], [5,7,5], [7,7,5],$ and $[5,9,5]$) besides our strategy (i.e., $[7,9,5]$) lead to a decrease in 
F1-ME(p) compared to F1-ME(0.5), and in certain situations, F1-ME(p) is even lower than F1-ME(1.0) on both datasets. This variance in results may be 
attributed primarily to our setting of values based on the actual distribution of the three categories (i.e., ME, MaE, and background). Therefore, we 
establish values in $h$ using a video-level distribution consistency strategy that can capture the majority of ME TP samples to boost the results of 
F1-ME(p) in the optimal proposal set, making it almost equivalent to F1-ME(0.5) in the overall proposal set, and significantly greater than F1-ME(1.0) 
in the overall proposal set on both datasets.

\textcolor{black}{Specially, from Table \ref{tab3}, we find that tuning the parameter corresponding to MaE in $h$ can affect the results of ME. The reason is that choosing 
a smaller value in $h$ results in a larger $k$, which makes more related snippets involved in model training in Equation \ref{equ3.7}. 
Furthermore, there is often a high probability for contextual background snippets to have relatively large logits \cite{fu2022compact, lee2021learning, huang2022weakly}, 
thus during testing, the durations of the generated proposals with smaller $h$ for training are generally longer than the durations of proposals with larger $h$ for training.
This leads to a higher chance to have smaller IoUs for the proposals that match the ME ground truth intervals. Next, we apply NMS to select proposals based on processed 
logits (in Section \ref{Section 3.5}). This operation will make some short proposals (potentially the ME proposals) be lost. 
Both of these factors have negative impacts on ME spotting.}
\begin{table}[htbp]
  \centering
  \small 
  \setlength\tabcolsep{10pt}
  \caption{Performances with various cross-dataset training strategies on the CAS(ME)$^2$ and SAMM-LV datasets.
  ``Separate1'' denotes CAS(ME)$^2$ for training and SAMM-LV for testing. 
  ``Separate2'' denotes SAMM-LV for training and CAS(ME)$^2$ for testing. 
  ``Merge'' signifies the merging of CAS(ME)$^2$ and SAMM-LV into a single dataset for both training and testing.}
    \begin{tabular}{c|cccccc}
    \toprule
    \multirow{2}[4]{*}{Measure} & \multicolumn{3}{c}{Cross-dataset Training Strategy} \\
    \cmidrule{2-4}          & Separate1 & Separate2 & Merge \\  
    \midrule
    F1-ME(0.5)  &0.086 &0.056 &0.088  \\
    F1-ME(1.0)  &0.071 &0.030 &0.055  \\
    F1-ME(p)    &0.019 &0.000 &0.014  \\                                                                                                                          
    Recall      &0.154 &0.130 &0.165  \\
    Precision   &0.288 &0.187 &0.187  \\
    F1-score    &0.201 &0.154 &0.175  \\
    \bottomrule
    \end{tabular}
  \label{tab8}
\end{table}

% \begin{table}[htbp]
%   \centering
%   \small 
%   \setlength\tabcolsep{10pt}
%   \caption{Performances with various truncation thresholds on the CAS(ME)$^2$ dataset.}
%     \begin{tabular}{c|cccccc}
%     \toprule
%     \multirow{2}[4]{*}{Measure} & \multicolumn{3}{c}{truncation Threshold} \\
%     \cmidrule{2-4}          & 0.05 & 0.1 & 0.2 \\  
%     \midrule
%     F1-ME(0.5)  &0.167 &0.167 &0.167 \\
%     F1-ME(1.0)  &0.108 &0.108 &0.108 \\
%     F1-ME(p)    &0.169 &0.169 &0.169 \\
%     Recall      &0.266 &0.266 &0.266 \\
%     Precision   &0.415 &0.415 &0.415 \\
%     F1-score    &0.324 &0.324 &0.324 \\
%     \bottomrule
%     \end{tabular}
%   \label{tab8.1}
% \end{table}

\subsubsection{Effect of Different Snippet Durations}
We investigate the effect of different snippet durations on model training. It is clear that the longer each snippet is, the fewer snippets are 
sampled from a video. In addition, because each snippet is represented as 1024-dimensional features extracted by the I3D model, fewer snippets 
in one video cannot learn fine-grained information, while too many may introduce too much noise. \textcolor{black}{Note that in order to avoid mutual interference 
between neighboring snippets, each video in this paper is segmented into a sequence of non-overlapping snippets.} Here, we test snippet durations 
of 4, 8, and 16 for CAS(ME)$^2$, and 16, 32, and 48 for SAMM-LV, respectively, during model training, and we keep the same settings during testing.

Table \ref{tab4} shows that the optimal snippet durations in terms of numbers of frames for CAS(ME)$^2$ and SAMM-LV are 8 and 32, respectively. A 
snippet duration that is either too long or too short reduces the spotting capability of our spotting model. This means that too short of a snippet 
duration may introduce too much noisy information, while a too long duration may lose sensitivity to short MEs. In particular, longer snippet 
durations greatly inhibit ME spotting on the two datasets. \textcolor{black}{Moreover}, on the CAS(ME)$^2$ dataset, the model fails to spot MEs when the duration is 16.
The above results suggest that the snippet duration should not be close to or exceed the maximum duration of the MEs in order to prevent loss of important information, 
nor be too small to avoid introducing excessive noise.

\subsubsection{Effect of Loss Function}
Each component of the loss function in Equation \ref{equ3.22} serves a specific role in refining our model. To assess the effectiveness of each 
component, we experiment with various combinations of them in the loss functions while keeping the same configuration. \textcolor{black}{We investigate the importance 
of five components: $\mathcal{L}_{sc}$, $\mathcal{L}_{dc}^1$, $\mathcal{L}_{dc}^2$, $\mathcal{L}_{dc}^3$, and $\mathcal{L}_{fc}$. Specifically, 
in comparison to CO$_2$-Net \cite{hong2021cross}, we introduce $\mathcal{L}_{sc}$ to enhance the inter-modal consistency and upgrade the MIL-Based losses 
to $\mathcal{L}_{dc}^1$ and $\mathcal{L}_{dc}^2$. In addition, we incorporate a video-level distribution consistency strategy with $\mathcal{L}_{dc}^3$ to 
focus on learning pure features of MaEs. At last, we substitute the existing co-activity similarity loss with our attention-guided feature consistency 
loss $\mathcal{L}_{fc}$.}

Table \ref{tab5} illustrates the impact of various loss components on the performance of MC-WES. \textcolor{black}{The results show that adding $\mathcal{L}_{sc}$ can 
significantly improve the performances of overall expression and ME spotting. Furthermore, to provide a more detailed comparison of
the differences among the three losses, we use different combinations of them to train our model. We find that the model performance is more improved by adding $\mathcal{L}_{gl}$ alone 
than by adding $\mathcal{L}_{sl}$ or $\mathcal{L}_{sc}$, and the model trained with any two of the three losses cannot outperform the model trained with $\mathcal{L}_{gl}$ alone. 
The reason is that there is the lack of MIL classification loss.} This highlights the effectiveness of our modal-level saliency consistency strategy with CSCM. 

Regarding MIL losses, we find that their inclusion in any combination 
enhances the model's spotting capability. Specifically, when MC-WES uses $\mathcal{L}_{sl}$, $\mathcal{L}_{gl}$ and $\mathcal{L}_{sc}$ with the extra 
addition of $\mathcal{L}_{dc}^1$ and $\mathcal{L}_{dc}^2$, the performance is improved most dramatically in recall, precision, and F1-score. This reinforces the validity 
of our video-level distribution consistency and label-level duration consistency strategies. 

\textcolor{black}{Furthermore, adding $\mathcal{L}_{dc}^3$ alone slightly improves MaE spotting in ``EXP 8'' and ``EXP 11''. 
In contrast, we introduce $\mathcal{L}_{dc}^3$ into our model that already uses $\mathcal{L}_{dc}^2$ and the results show a slight decrease in performance in ``EXP 10'' and ``EXP 12''. }
We also integrate $\mathcal{L}_{fc}$ into MC-WES, and the results in 
Table \ref{tab5} demonstrate that $\mathcal{L}_{fc}$ enhances the model's spotting ability. This confirms the effectiveness of 
our proposed segment-level feature consistency strategy. 

However, when all the aforementioned losses are employed except for $\mathcal{L}_{dc}^2$, there is a significant 
decrease in the effectiveness in ``EXP 12'' and ``EXP 15''. One potential reason could be that $\mathcal{L}_{fc}$ relies on representational foreground 
features to learn similarities, whereas the lack of $\mathcal{L}_{dc}^2$ prevents the model from learning good representational foreground features. 

\textcolor{black}{In the case of ME spotting, adding $\mathcal{L}_{sc}$ or $\mathcal{L}_{fc}$ does not significantly enhance ME spotting and adding $\mathcal{L}_{dc}^1$, $\mathcal{L}_{dc}^2$
or $\mathcal{L}_{dc}^2$ can lead to improvements. 
Remarkably, using all five components together noticeably enhances the performance of ME spotting. 
This underscores the effectiveness of our proposed strategies. }

\begin{table*}[htbp]
  \centering
  \small 
  \setlength\tabcolsep{5pt}
  \caption{Comparison with state-of-the-art models on the CAS(ME)$^2$ dataset. The numbers in bold highlight the best values among the compared fully- or weakly-supervised methods.}
    \begin{tabular}{c|l|cccccc}
    \toprule
    Supervision & Method &F1-ME(0.5)&F1-ME(1.0)&F1-ME(p)&Recall&Precision&F1-score \\
    \midrule
    \multirow{8}[2]{*}{Full}  &He et al. (2020)\cite{he2020spotting}      &   -  &   -  &0.008 &0.020 &0.364 &0.038 \\
                              &Zhang et al. (2020)\cite{zhang2020spatio}  &   -  &   -  &0.055 &0.085 &0.406 &0.140 \\
                              &MESNet (2021) \cite{wang2021mesnet}        &   -  &   -  &  -   &  -   &   -  &0.036 \\
                              &Yap et al. (2021) \cite{yap20213dcnn}      &   -  &   -  &0.012 &  -   &   -  &0.030 \\
                              &LSSNet (2021) \cite{yu2021lssnet}          &   -  &   -  &0.063 &  -   &   -  &0.327 \\
                              &He et al. (2021) \cite{yuhong2021research} &   -  &   -  &\textbf{0.197} &  -   &   -  &0.343 \\
                              &MTSN (2022) \cite{liong2022mtsn}           &   -  &   -  &0.081 &0.342 &0.385 &0.362 \\
                              &Zhao et al. (2022) \cite{zhao2022rethink}  &   -  &   -  &   -  &   -  &   -  &0.403 \\
                              &LGSNet (2023)\cite{yu2023lgsnet}           &   -  &   -  &   -  &\textbf{0.367} &\textbf{0.630} &\textbf{0.464} \\

    \midrule

    \multirow{4}[2]{*}{Weak}  &HAM-Net (2021) \cite{islam2021hybrid}     &0.010 &0.007 &0.000 &0.042 &0.090 &0.057 \\
                              &CO$_2$-Net (2021) \cite{hong2021cross}    &0.057 &0.031 &0.000 &0.095 &0.153 &0.117 \\
                              &FTCL (2022) \cite{gao2022fine}            &0.092 &0.022 &0.000 &0.048 &0.070 &0.057 \\                                
                              &\textbf{MC-WES}                           &\textbf{0.167} &\textbf{0.108} &\textbf{0.169} &\textbf{0.266} &\textbf{0.415} &\textbf{0.324} \\

    \bottomrule
    \end{tabular}
  \label{tab9}
\end{table*}

\begin{table*}[htbp]
  \centering
  \small 
  \setlength\tabcolsep{5pt}
  \caption{Comparison with state-of-the-art models on the SAMM-LV dataset.}
    \begin{tabular}{c|l|cccccc}
    \toprule
    Supervision & Method &F1-ME(0.5)&F1-ME(1.0)&F1-ME(p)&Recall&Precision&F1-score \\
    \midrule
    \multirow{8}[2]{*}{Full}  &He et al. (2020) \cite{he2020spotting}     &   -  &   -  &0.036 &0.029 &0.101 &0.045 \\
                              &Zhang et al. (2020) \cite{zhang2020spatio} &   -  &   -  &0.073 &0.079 &0.136 &0.100 \\
                              &MESNet (2021) \cite{wang2021mesnet}        &   -  &   -  &   -  &   -  &  -   &0.088 \\
                              &Yap et al. (2021) \cite{yap20213dcnn}      &   -  &   -  &0.044 &   -  &  -   &0.119 \\
                              &LSSNet (2021) \cite{yu2021lssnet}          &   -  &   -  &\textbf{0.218} &   -  &  -   &0.290 \\
                              &He et al. (2021) \cite{yuhong2021research} &   -  &   -  &0.216 &   -  &  -   &0.364 \\
                              &MTSN (2022) \cite{liong2022mtsn}           &   -  &   -  &0.088 &0.260 &0.319 &0.287 \\
                              &Zhao et al. (2022) \cite{zhao2022rethink}  &   -  &   -  &   -  &   -  &   -  &0.386 \\
                              &LGSNet (2023)\cite{yu2023lgsnet}           &   -  &   -  &   -  &\textbf{0.355} &\textbf{0.429} &\textbf{0.388} \\
    \midrule
    \multirow{4}[2]{*}{Weak}  &HAM-Net (2021) \cite{islam2021hybrid}  &0.113 &\textbf{0.060} &0.028 &0.150 &0.113 &0.129\\
                              &CO$_2$-Net (2021) \cite{hong2021cross} &0.111 &0.058 &0.039 &0.230 &0.148 &0.181 \\
                              &FTCL (2022) \cite{gao2022fine}         &0.116 &0.048 &0.004 &0.142 &0.138 &0.140 \\ 
                              &\textbf{MC-WES}                        &\textbf{0.135} &0.055 &\textbf{0.135} &\textbf{0.263} &\textbf{0.178} &\textbf{0.212} \\

    \bottomrule
    \end{tabular}
  \label{tab10}
\end{table*}

\begin{table*}[htbp]
  \centering
  \small 
  \setlength\tabcolsep{5pt}
  \caption{Performances on the CAS(ME)$^3$ dataset.}
    \begin{tabular}{c|l|cccccc}
    \toprule
    Supervision & Method &F1-ME(0.5)&F1-ME(1.0)&F1-ME(p)&Recall&Precision&F1-score \\
    \midrule
    \multirow{3}[2]{*}{Full}  &SP-FD (2020)\cite{zhang2020spatio}      &0.010 &0.010 &  -   &  -   &  -   &  -  \\
                              &LSSNet (2021)\cite{yu2021lssnet}        &0.065 &0.065 &  -   &  -   &   -  &  -  \\
                              &LGSNet (2023)\cite{yu2023lgsnet}        &\textbf{0.171} &\textbf{0.136} &\textbf{0.099} &\textbf{0.292} &\textbf{0.196} &\textbf{0.235} \\

    \midrule
    \multirow{4}[2]{*}{Weak} &HAM-Net (2021)\cite{islam2021hybrid}     &0.008 &0.006 &0.000 &0.098 &0.030 &0.046 \\
                             &CO$_2$-Net (2021)\cite{hong2021cross}    &0.037 &0.018 &0.000 &0.118 &0.050 &0.070 \\
                             &FTCL (2022) \cite{gao2022fine}           &0.014 &0.012 &0.000 &0.106 &0.034 &0.052 \\
                             &\textbf{MC-WES}                          &\textbf{0.048} &\textbf{0.022} &0.000 &\textbf{0.141} &\textbf{0.060} &\textbf{0.084} \\
    \bottomrule
    \end{tabular}
  \label{tab11}
\end{table*}

\subsubsection{Effect of Different Post-processing}
\label{Section 4.4.5}
As mentioned in Section \ref{Section 3.5}, previous works \cite{islam2021hybrid, hong2021cross} has generally employed a multi-threshold method based on 
attention scores to select snippets 
for generating proposals, which are then used to calculate the mean average precision (mAP). To further assess the localization capability of the
model, the TAL task tends to calculate mAP under various intersection over union (IoU) thresholds. These results on mAP can adequately reflect the 
impact of different confidence counterparts on the proposals. However, the expression spotting task favors the use of a certain IoU threshold to 
filter proposals and calculate metrics. Once we use a multi-threshold approach to filter attention scores and generate proposals, a large number of 
negative samples will be produced.

To resolve this problem, our model employs a multi-top method to select a restricted number of snippets with high attention scores for proposal generation,
as discussed in Section \ref{Section 3.5}. Considering that the videos in the CAS(ME)$^2$ and SAMM-LV datasets vary in duration, we utilize 
different integers set in $M'$ for the two datasets. Subsequently, the category scores for these proposals are calculated according to the procedure described in 
Section \ref{Section 3.5}. \textcolor{black}{Because $h$ is predefined as $[7,9,5]$ for CAS(ME)$^2$ in video-level distribution consistency strategy in Section \ref{Section 3.3} 
during training, we set the start integer in the set $M'$ as 8 for a compromise between MEs and MaEs during testing. As the duration of the snippets on 
the SAMM-LV dataset is shorter than that on the CAS(ME)$^2$ dataset, to spot more proposals, we set the start integer in $M'$ as 2 for SAMM-LV.}

The results of Table \ref{tab7} show that our multi-top method is better than the multi-threshold approach in terms of recall, precision, and F1-score. 
This indicates that our model can spot more accurate snippets and produce fewer negative samples. The multi-threshold approach primarily shows a declining 
precision in addition to reducing the recall due to the large number of negative samples generated during testing. As for ME spotting, the performance reduction 
by the multi-threshold approach is remarkable on both datasets. On the CAS(ME)$^2$ dataset, F1-ME(P) is 0.028, which is close to zero, suggesting that most MEs 
fail to be spotted using this approach on this dataset.

\subsubsection{Cross-dataset Validation}
\label{Section 4.4.6}
\textcolor{black}{To evaluate the generalization of our proposed MC-WES across different datasets, we employ three cross-dataset training strategies: training with 
the CAS(ME)$^2$ dataset and testing with the SAMM-LV dataset, training with the SAMM-LV dataset and testing with the CAS(ME)$^2$ dataset, and the 
fusion of the CAS(ME)$^2$ and SAMM-LV datasets into a single dataset for both training and testing with LOSO learning strategy. These strategies are denoted in Table \ref{tab8} 
as ``separate1'', ``separate2'', and ``merge'', respectively.}

\textcolor{black}{The results in Table \ref{tab8} show that when our model is implemented with the ``separate1'' strategy, the results in terms of recall, precision and 
F1-score decrease by 2.4\%, 10.1\%, and 4.7\%, respectively, compared to the ``separate2'' strategy. This suggests the presence of significant distribution 
differences between the two datasets. As explained in Section \ref{Section 4.1}, there are two reasons for these differences. First, they arise from variations in 
sample density and duration. Second, the distinct labeling principles of the two datasets may introduce video-level noisy labels from SAMM-LV for training our weakly-supervised model.}

\textcolor{black}{The results reported in Table \ref{tab8} demonstrate that the F1-score achieved with the ``merge'' strategy is higher than that with the ``separate2'' strategy but 
lower than that with the ``separate1'' strategy. In addition, our model trained with the ``merge'' strategy yields best recall, but lower precision compared 
to the model learned with the ``separate1'' or ``separate2'' strategy. This suggests that increasing the number of training samples benefits our model in detecting 
more TP samples. However, the potential video-level noisy labels mentioned above may hinder our model's ability to reduce FP samples.}

\textcolor{black}{As for ME spotting, we can observe the same results as the overall expression spotting. This also reveals that insufficiently fine-grained labeling also affects ME spotting.}

\subsection{Comparison with State-of-the-art Methods}
To the best of our knowledge, MC-WES represents the first attempt to achieve frame-level expression spotting using weakly-supervised video-level labels. Therefore, 
we evaluate its performance by comparing it with recent fully-supervised state-of-the-art methods on the CAS(ME)$^2$ and SAMM-LV datasets. \textcolor{black}{Additionally, we assess MC-WES by comparing 
it with other recent weakly-supervised methods that are originally designed for WTAL task on the same datasets.}

Table \ref{tab9} shows that our proposed weakly-supervised MC-WES can achieve results that are somewhat comparable to the representative fully-supervised methods on the CAS(ME)$^2$ 
dataset. There is not much performance degradation in the spotting of MEs. Compared with MTSN \cite{liong2022mtsn}, MC-WES obtains an improved precision rate on the CAS(ME)$^2$ dataset. 
Table \ref{tab10} indicates that our method also achieves acceptable results on the SAMM-LV dataset. \textcolor{black}{Compared with MTSN\cite{liong2022mtsn}, MC-WES needs improvement in terms of the 
precision rate on the SAMM-LV dataset.} Notably, in Section \ref{Section 4.1}, 
we discuss the challenges posed by the limited ground truth intervals in SAMM-LV, which were not filtered for long-tail intervals, potentially affecting the results.

Furthermore, \textcolor{black}{Tables \ref{tab9} and \ref{tab10} indicate that our MC-WES remarkably outperforms other weakly-supervised methods in terms of recall, precision, 
and F1-score on both the CAS(ME)$^2$ and SAMM-LV datasets, indicating its effectiveness. As for the case of ME spotting by the weakly-supervised methods, our model performs 
clearly best in terms of F1-ME(0.5), F1-ME(P), and F1-ME(1.0) on the CAS(ME)$^2$ dataset, and leads largely in F1-ME(0.5) and F1-ME(P) on the SAMM-LV dataset.}

\textcolor{black}{Considering that CAS(ME)$^2$ and SAMM-LV contain a very limited number of samples compared with the datasets used in other computer vision fields,
we further conduct evaluation on a relatively large dataset--CAS(ME)$^3$, which contains 956 videos.}

\textcolor{black}{Table \ref{tab11} presents the superior performance of MC-WES compared to recent weakly-supervised methods that are originally designed for WTAL task, as evidenced by the
noteworthy improvements across multiple metrics. 
Specifically, MC-WES achieves a minimum 1.4\% enhancement in F1-score, a minimum 2.3\% increase in recall, and a 1\% rise in precision. These results emphasize the substantial 
progress achieved by our method on larger datasets. In terms of ME spotting, while F1-ME(P) keeps consistent, both F1-ME(0.5) and F1-ME(1.0) exhibit significant enhancements, 
confirming the effectiveness of our approach.}

\section{Conclusion}
In this paper, to avoid the requirement of tedious frame-level labeling for the ME datasets, we explored the use of a weakly-supervised video-level MIL-based framework named MC-WES. 
This approach aims to spot frame-level expressions through the integration of multi-consistency collaborative mechanisms, which encompass strategies such as modal-level saliency 
consistency, video-level distribution consistency, label-level duration consistency, and segment-level feature consistency. Specifically, The modal-level saliency consistency strategy 
is utilized to capture the key correlations between raw images and optical flow. Furthermore, the video-level distribution consistency strategy merges information of different 
sparseness in the sample distribution, and the label-level duration consistency strategy exploits the difference in duration of facial muscles. To learn more representational 
features and mitigate the discrepancy between classification and localization, we employ the segment-level feature consistency strategy. Extensive experiments on the CAS(ME)$^2$, 
CAS(ME)$^3$, and SAMM-LV datasets are conducted to validate MC-WES. The results demonstrate that the proposed multi-consistency collaborative mechanism enables our weakly-supervised 
spotting method to achieve results comparable to those of fully-supervised spotting methods and outperforms other weakly-supervised methods.

Although the MC-WES framework relies on the outer-inner scores from Section \ref{Section 3.5} to select proposals and then calculate precision and recall rates, we believe that mAP 
is a more appropriate metric to evaluate the spotting capability of the model. As an important future work, we plan to develop a more refined framework to enhance the model's robustness 
of expression spotting when there exist a possible bias in labeling and large-scale duration of ground truth intervals inevitably, such as those on the SAMM-LV dataset. Furthermore, 
\textcolor{black}{Lu et al. \cite{lu2022more} have attempted to quantify the intensity of facial expressions using electromyography (EMG) signals. This certainly inspires us to investigate 
the spotting of MEs and MaEs based on the intensity of facial movements.}

% or


\ifCLASSOPTIONcompsoc
  \section*{Acknowledgments}
\else
  \section*{Acknowledgment}
\fi

This work was supported by STI2030-Major Projects (\#2022ZD0204600), Sichuan Science and Technology Program (2022ZYD0112) 
and the Natural Science Foundation of Sichuan Province (2023NSFSC0640).

\ifCLASSOPTIONcaptionsoff
  \newpage
\fi



%\bibliographystyle{IEEEtran}


% \end{thebibliography}

\bibliographystyle{IEEEtran}
\bibliography{ref}


\begin{IEEEbiography}[{\includegraphics[width=1in,height=1.25in,clip,keepaspectratio]{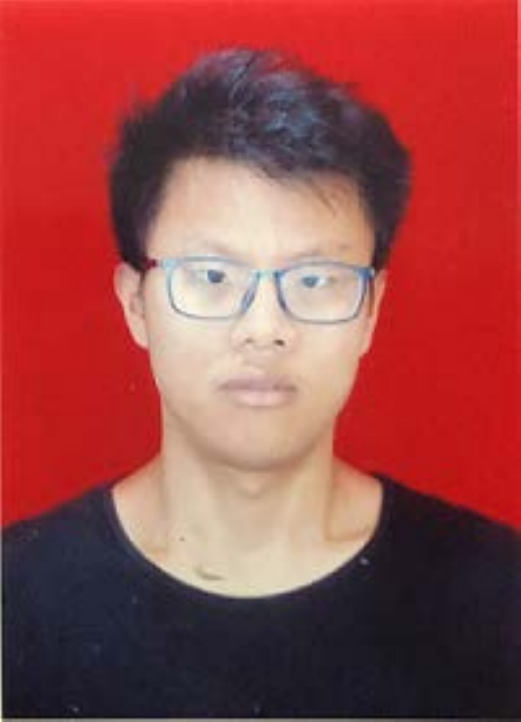}}]
  {Wang-Wang Yu} received the M.S. degree in biomedical engineering from University of Electronic Science and Technology of China (UESTC) in 2020. 
  He is now pursuing his Ph.D. degree in UESTC. His research interests include video understanding, emotional analysis, weakly-supervised learning.
\end{IEEEbiography}

\begin{IEEEbiography}[{\includegraphics[width=1in,height=1.25in,clip,keepaspectratio]{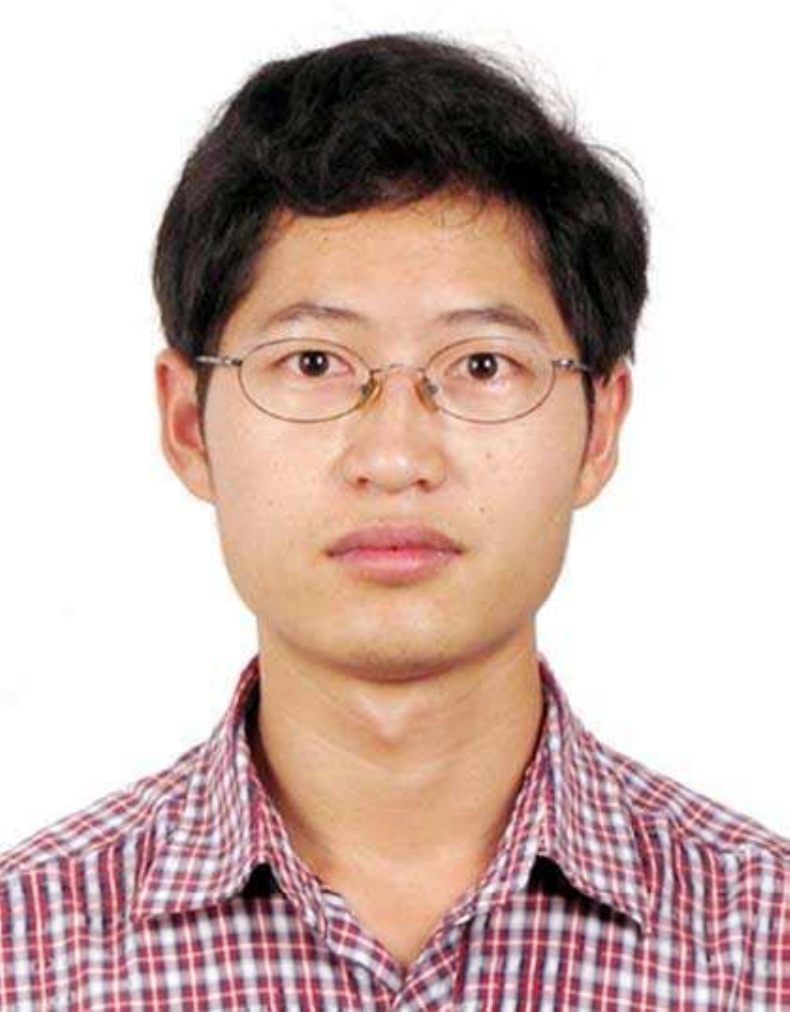}}]
  {Kai-Fu Yang} received the Ph.D. degree in biomedical engineering from the University of Electronic Science and 
  Technology of China (UESTC), Chengdu, China, in 2016. He is currently an associate research professor with 
  the MOE Key Lab for NeuroInformation, School of Life Science and Technology, UESTC, Chengdu, China. His research 
  interests include cognitive computing and brain-inspired computer vision.
\end{IEEEbiography}

\begin{IEEEbiography}[{\includegraphics[width=1in,height=1.25in,clip,keepaspectratio]{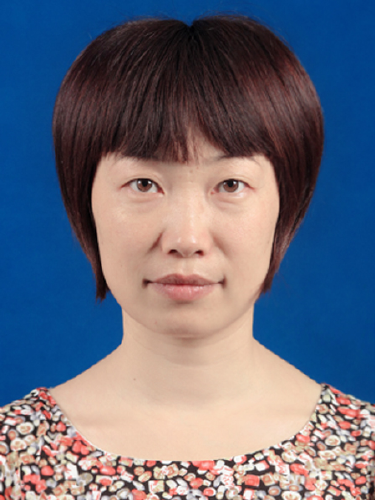}}]
  {Hong-Mei Yan} received the Ph.D. degree in biomedical engineering from Chongqing University in 2003. She is now a Professor 
  with the MOE Key Laboratory for NeuroInformation, University of Electronic Science and Technology of China, Chengdu, China. 
  Her research interests include visual cognition, visual attention, visual encoding and decoding.
\end{IEEEbiography}

\begin{IEEEbiography}[{\includegraphics[width=1in,height=1.25in,clip,keepaspectratio]{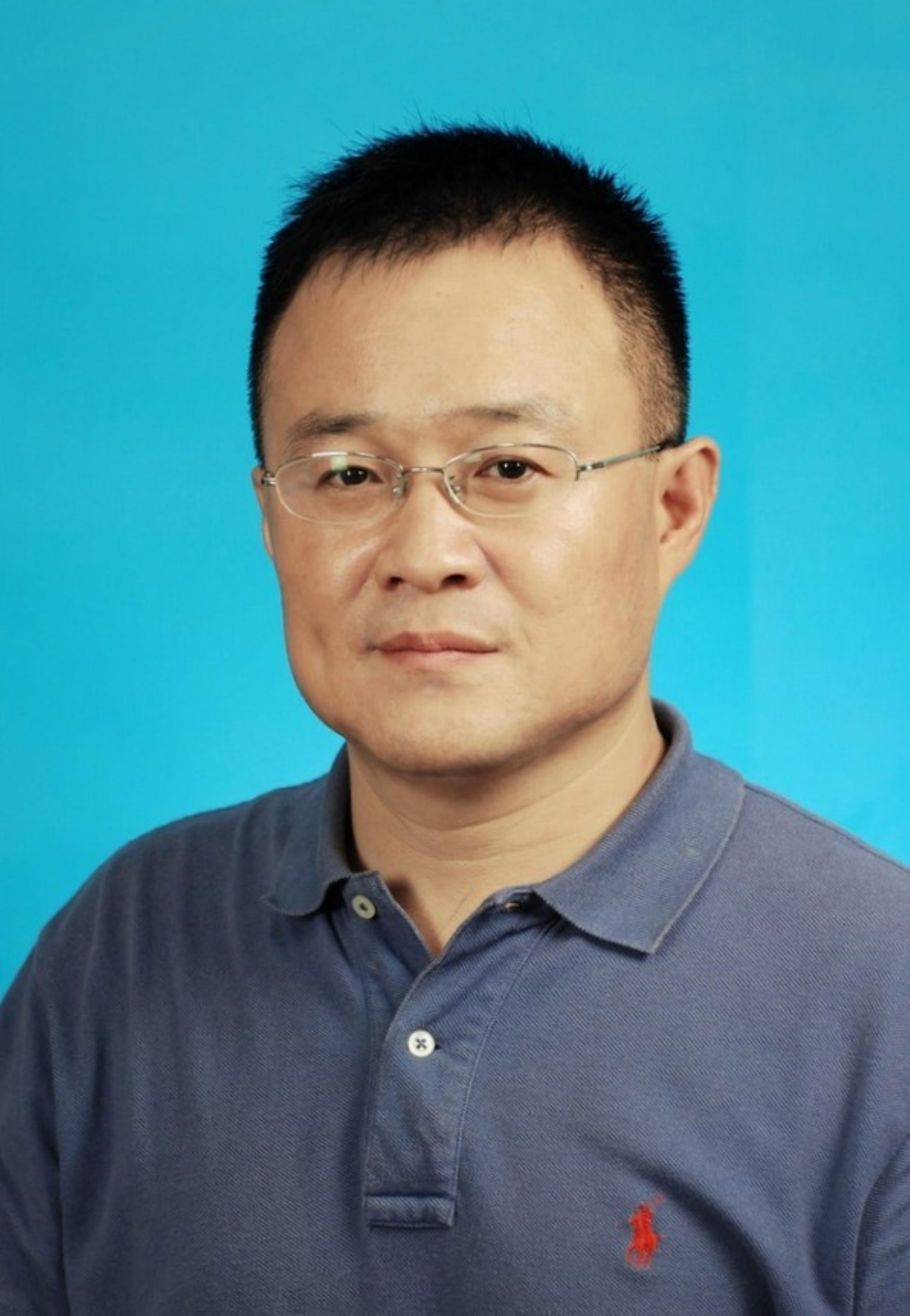}}]
  {Yong-Jie Li} (Senior Member, IEEE) received the Ph.D. degree in biomedical engineering from
  UESTC, in 2004. He is currently a Professor with the Key Laboratory for NeuroInformation of Ministry of
  Education, School of Life Science and Technology, University of Electronic Science and Technology of
  China. His research focuses on building of biologically inspired computational models of visual perception and 
  the applications in image processing and computer vision.
\end{IEEEbiography}


%\vfill


\end{document}